\documentclass[11pt]{article}

\usepackage{acl}
\usepackage{times}
\usepackage{latexsym}
\usepackage[T1]{fontenc}
\usepackage[utf8]{inputenc}
\usepackage{microtype}
\usepackage{inconsolata}
\usepackage{graphicx}
\usepackage{multirow}
\usepackage{amsmath}
\usepackage{booktabs}
\usepackage{listings}
\usepackage{amssymb}
\usepackage{xcolor}
\usepackage{url}
\usepackage{enumitem}
\usepackage{array}
\usepackage{fontawesome}
\usepackage{siunitx}
\usepackage{bbm}

\setlist[itemize]{leftmargin=*,topsep=2pt,itemsep=1pt,parsep=0pt}
\setlist[enumerate]{leftmargin=*,topsep=2pt,itemsep=1pt,parsep=0pt}

\lstdefinestyle{promptstyle}{
  basicstyle=\ttfamily\scriptsize,
  breaklines=true,
  breakatwhitespace=true,
  columns=fullflexible,
  frame=single,
  framesep=4pt,
  xleftmargin=4pt,
  captionpos=b,
}

\newcommand{\addrrel}{\texorpdfstring{\mbox{\textsc{Address}}}{Address}}
\newcommand{\broadrel}{\texorpdfstring{\mbox{\textsc{Broaden}}}{Broaden}}
\newcommand{\specrel}{\texorpdfstring{\mbox{\textsc{Specify}}}{Specify}}
\newcommand{\system}{\texorpdfstring{\mbox{\textsc{Ratio}}}{Ratio}}

\title{\system{}: A Benchmark for Retrieval Across Typed Ideation Operations in Scientific Literature}

\author{Maayan Sharon \\
  The Hebrew University of Jerusalem \\
\\\And
  Tom Hope \\
  The Hebrew University of Jerusalem \\
  The Allen Institute for AI (AI2)
\AND
  {\normalfont\raisebox{1em}[0pt][0pt]{
  \faGlobe~\href{https://maayansharon10.github.io/RATIO/}{Project} \quad
  \faGithub~\href{https://github.com/maayansharon10/RATIO}{Github} \quad
  \faDatabase~\href{https://huggingface.co/collections/maayans/ratio}{Data \& Models}}}
}

\begin{document}
\maketitle

\begin{abstract}
Retrieved scientific literature can serve as inspiration for both human and AI scientists. Inspiration can take different forms: prior work may directly suggest how to address a problem, or surface directions at different levels of abstraction---zooming out to a more general view or zooming in to a concrete realization. We introduce
\system{} (Retrieval Across Typed Ideation Operations), a large-scale benchmark in which relevance is defined
by three operations which we name \emph{ideation moves}: \addrrel{} retrieves potential approaches for
stated problems, \broadrel{} retrieves more general
formulations, and \specrel{} retrieves concrete instantiations. \system{} is
constructed from millions of full-text scientific papers across CS literature via a general recipe that extends discourse-marker distant supervision---previously used only for classification---to corpus-scale retrieval, combined with extensive LLM and human vetting. 
Experiments show that operation-specific fine-tuning substantially boosts retrievers and LLM reranking helps further, but much room for improvement remains. Researchers blind to condition found the fine-tuned retriever's candidates more useful as inspirations.
\system{} provides a scalable training and evaluation framework for retrieval components that support literature-grounded ideation, opening up new research avenues in scientific inspiration retrieval. We release full data, code, models, and annotations.
\end{abstract}

\begin{figure}[t]
    \centering
    \includegraphics[width=1\linewidth]{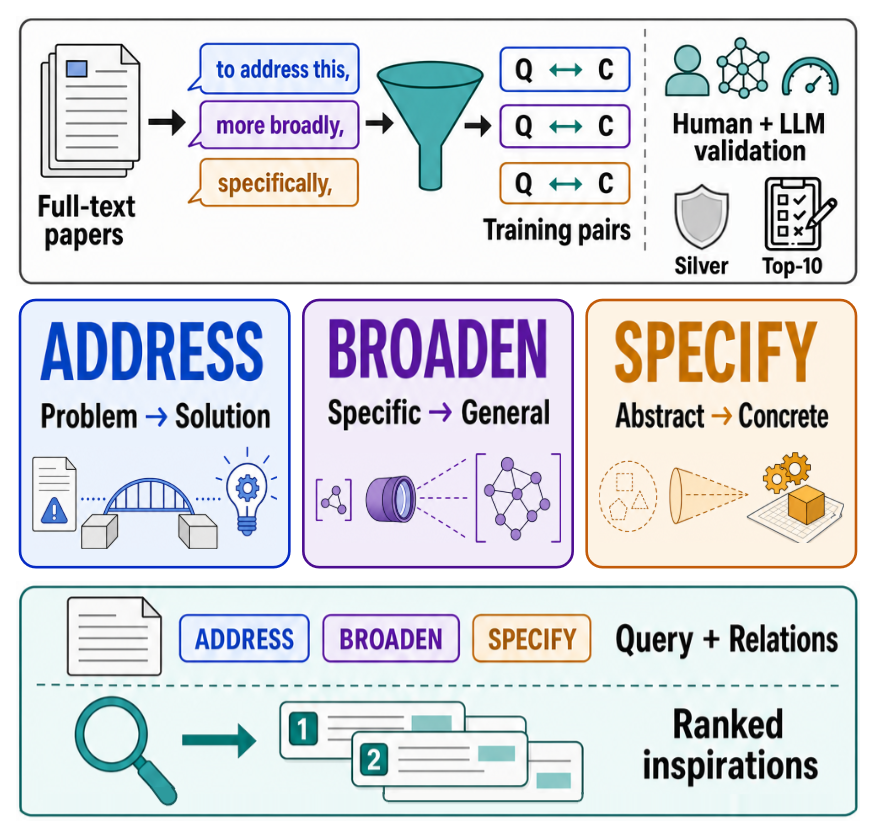}
    \caption{We construct \system{} from full-text scientific papers using discourse-marker supervision and human–LLM validation to support retrieval across three ideation moves:  \addrrel, \broadrel, \specrel.}
    \label{fig:figure1}
\end{figure}
\section{Introduction}
\label{sec:intro}

Engaging with prior work is central to how researchers develop new ideas: the literature can spark analogical transfer, conceptual recombination, and problem reframing \cite{sweed-etal-2025-muse,sternlicht-hope-2026-chimera}. Consider for example a researcher grappling with reward hacking in RL-trained LLMs. Three different inspirations may be helpful: one proposing a method (e.g., reward-model ensembles), one zooming out to a more general principle (e.g., Goodhart's law), and one zooming in on a concrete instantiation of that principle in a new setting (e.g., recommender systems gaming engagement metrics) to provide an inspiration from another problem. Each one enables a distinct move through the ideation space---addressing a problem, abstracting it, or instantiating it.

This distinction matters for systems that assist scientific
research with literature-grounded hypothesis
generation
\citep{wang-etal-2024-scimon,garikaparthi-etal-2025-mir,radensky2026scideator,
vasu-etal-2025-hyper,liu-etal-2026-researchbench,gu-etal-2026-mori}. In this
setting, retrieval determines which prior mechanisms, abstractions, and problem
formulations are available as inspirations to move the ideation process forward. 
For human researchers and AI agents, retrieval helps define the reachable ideation space; showing related mechanisms at different abstraction levels as inspirational stimuli is known to boost ideation \cite{Hope2021ScalingCI, sweed-etal-2025-muse}.

Most scholarly retrieval benchmarks, however, evaluate topical relevance or
whether a document answers a search query \citep{ajith-etal-2024-litsearch}.
These criteria are necessary for literature search but do not distinguish
between passages that play different ideation roles.  
We introduce \system{} (Retrieval Across Typed Ideation Operations), a benchmark for scientific inspiration retrieval (Figure~\ref{fig:figure1}). Grounded in cognitive theories of ideation, we view external inspirations as possible moves through an ideation space \citep{dorst2001creativity,goldschmidt2014linkography}. Under this view, a retrieved statement is useful when it suggests a particular next move. \system{} organizes retrieval into three moves or operation-specific tasks that capture complementary forms of inspiration. \addrrel{}
retrieves an approach or insight that responds to a problem in the query. \broadrel{} retrieves a formulation
at a broader scope or greater generality. \specrel{} retrieves a concrete
instance or narrower formulation. Within each task, relevance is determined by latent relational patterns with often limited surface expression, making the desired query–candidate relation difficult for retrieval systems to recover \cite{tchuindjo2026obliq}. 

Constructing such a benchmark at the scale required for retriever training and evaluation is
challenging. We propose a scalable distant supervision approach that uses
explicit high-precision discourse markers in full-text papers. Phrases such as \emph{to
address this problem}, \emph{more generally}, and \emph{as an example} identify adjacent
sentences that authors themselves present as problem-addressing, broader, or
more specific. Unlike prior discourse-marker distant supervision, which is confined to sentence-pair classification \cite{sadat-caragea-2022-scinli,sileo2019mining}, we extend markers to define relation-conditioned ideation retrieval over a corpus of millions of candidates (§\ref{sec:related}).

\system{} is constructed through a multi-stage supervision and validation pipeline. We first build move-specific lexicons of discourse cues for \addrrel, \broadrel, and \specrel, combining manual corpus analysis, rule-based expansion, and generation by multiple LLMs. Candidate cues are filtered through expert review and LLM validation before being used to mine  sentence pairs from full-text papers. We further create a higher-precision silver evaluation set, using human vetting of discourse markers and human-calibrated LLM judgments. Finally, we also validate top-10 retrieved candidates to account for false negatives. 

Experiments with retrievers show that the proposed operations are not captured reliably by models (including reasoning-trained and late-interaction models), and operation fine-tuning yields improvements for all three operations, while absolute performance remains limited. In a human evaluation, researchers blind to condition found substantially more useful inspirations in the retriever's top 10 after fine-tuning than before.

\noindent\textbf{Our contributions:}
\begin{itemize}[leftmargin=0pt]
    \item We define \emph{scientific ideation-move inspiration retrieval} tasks,
    where relevance is specified by an ideation operation:
    addressing a problem, broadening a formulation, or providing a concrete instantiation.
    \item  We introduce a general methodology for constructing relation-conditioned retrieval benchmarks from raw corpora via discourse-marker supervision---including a multi-stage construction pipeline and human-calibrated LLM validation---extending marker-based distant supervision from its traditional classification setting to retrieval.
    \item We instantiate this methodology as \system{}, a large-scale full-text benchmark for the three ideation operations, with temporally held-out, high-quality evaluation sets with human vetting. We benchmark retrievers, showing that
    relation-specific fine-tuning substantially improves results and finds useful cross-paper inspiration candidates while leaving a considerable performance gap. In a human evaluation, researchers blind to condition prefer the fine-tuned retriever's lists; we will release over 2,000 human query--candidate annotations and list-level annotations for 125 queries.
\end{itemize}

\begin{figure*}[t]
   \centering \includegraphics[width=1\textwidth]
{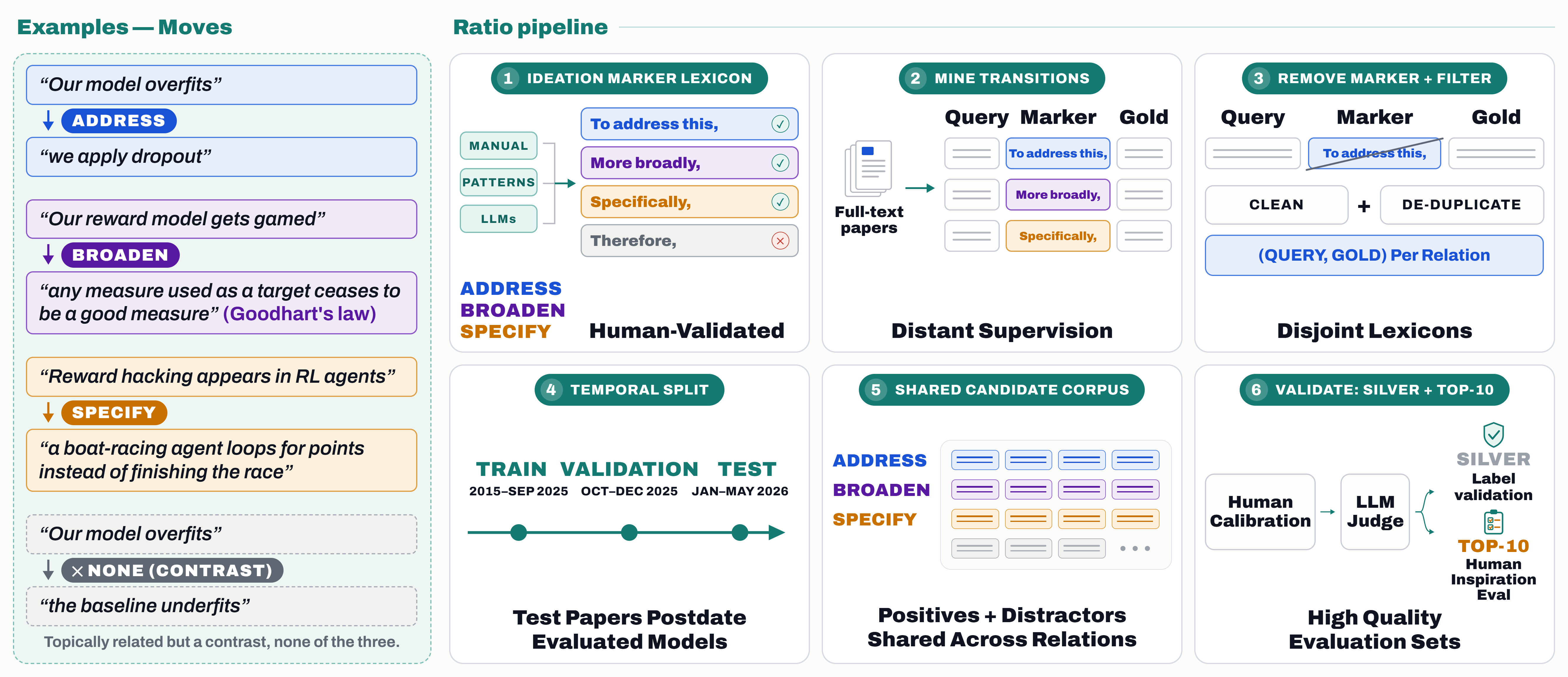}
    \caption{Left: one example per ideation move, and a topically related sentence that realizes none of the moves.
    Right: construction of \system. Validated discourse markers identify pairs of consecutive sentences in full-text papers that realize \addrrel, \broadrel, or \specrel. In each pair, the first sentence becomes the query, and the second, without its marker, becomes the query's gold sentence (its correct answer). Publication dates define the data splits; within a split, all three moves share one pool of candidate sentences. An LLM judge calibrated against human labels filters the test pairs into the silver test set (\S\ref{sec:silver_set_construction}). Researchers evaluate the top 10 retrieved candidates (\S\ref{sec:human_inspiration}).}
    \label{fig:pipeline}
\end{figure*}

\section{Related Work}
\label{sec:related}

\textbf{Literature-grounded scientific ideation.}
Recent work uses the scientific literature to support hypothesis generation and research planning \cite{wang-etal-2024-scimon,Hu2025NOVAAI,vasu-etal-2025-hyper,liu-etal-2026-researchbench,gu-etal-2026-mori}. These works evaluate generation or reasoning conditioned on scientific evidence, whereas \system{} evaluates the retrieval step that determines which inspirations such systems see.
\citet{Hope2021ScalingCI} and \citet{sweed-etal-2025-muse} represent problems, mechanisms, and abstractions as functional concept graphs, in which a problem can be linked to mechanisms from other domains that may address it. CHIMERA mines and retrieves naturally occurring recombinations of ideas \citep{sternlicht-hope-2026-chimera}. These works share our view that scientific relevance is not limited to topical similarity. \system{} differs from them in being an automatically constructed benchmark for sentence-level retrieval in which the requested move is given explicitly.
\citet{garikaparthi-etal-2025-mir} introduce methodology inspiration retrieval (MIR), a paper-level task defined primarily over abstracts, in which queries are research proposals and relevance labels are derived from citations that indicate methodological lineage. \system{} instead retrieves sentences from full texts and distinguishes three types of inspiration.

\textbf{Scholarly retrieval benchmarks.}
Benchmarks such as LitSearch evaluate whether retrieved papers answer literature-search questions \citep{ajith-etal-2024-litsearch}, which mainly measures topical or informational relevance. In \system{}, relevance instead depends on the role that a candidate plays relative to the query (\S\ref{sec:formulation}), and the relevance labels come from discourse markers, not from manual annotation, citations, or search logs. Work on reasoning-intensive retrieval \citep{su2025bright,shao2025reasonir} shares our premise that relevance can require inference beyond topical similarity, and we include retrievers trained for it in our evaluation (\S\ref{sec:methods}).

\textbf{Scientific sentence relations.}
SciNLI is a natural language inference (NLI) dataset for scientific text, and MSciNLI extends it to multiple scientific domains \citep{sadat-caragea-2022-scinli,sadat-caragea-2024-mscinli}. Both assign given sentence pairs to entailment-style classes. They do not evaluate retrieval from a large corpus, and their label sets do not distinguish our moves: helping to address a problem, or stating a more general or more specific version of another sentence. Both datasets are also built from a few discourse markers, whereas our lexicons contain 4,252 markers, 809 of which occur in our corpus.

\textbf{Discourse markers as distant supervision.}
Discourse markers have also served as distant supervision outside scientific NLI. For example, \citet{sileo2019mining} learn sentence representations by predicting the marker that links two sentences, or its category, among 174 markers mined from web text. In this line of work, the marker is the prediction target of a sentence-pair task. In \system{}, by contrast, the marker defines relevance. The retriever is given the move that the marker signals and must find the marked sentence in a large pool.

\section{Problem Definition}
\label{sec:formulation}
We formulate the task of scientific ideation-move inspiration retrieval. Given a scientific statement and a desired move through the ideation space, the task is to retrieve a candidate inspiration statement corresponding to that move.

Formally, a {\textit{query}} $q$ may be a sentence describing a scientific research problem,
claim, observation, method, or methodological objective; a {\textit{relation}} $r \in \mathcal{R} = \{\addrrel, \broadrel, \specrel\}$ is the discourse
relation that must hold between two sentences; and a {\textit{scientific
literature corpus}} $\mathcal{C} = \{c_1, \ldots, c_N\}$ is the set of all
candidate sentences from scientific papers. Given $q$ and $r$, the task is to produce a ranking under the relevance criterion associated with $r$
\begin{equation}
    \widehat{\mathcal{C}}^{\,k}_{r}(q)
    = \bigl\langle \hat{c}_1, \ldots, \hat{c}_k \bigr\rangle
\end{equation}
of the top-$k$ sentences in $\mathcal{C}$ by how well each candidate
$c$ instantiates $r$ with respect to $q$. 

We define the relations as follows,
for a query $q$ and a candidate $c \in \mathcal{C}$:
\begin{itemize}[nosep, leftmargin=1.2em]
    \item \textbf{\addrrel:} $q$ articulates a problem, and $c$ an approach or insight
    that can help address it.
    \item \textbf{\broadrel:} $c$ reformulates $q$ at a broader scope or
    greater generality, such that $q$ can be viewed as a particular instance
    or manifestation of $c$.
    \item \textbf{\specrel:} $c$ instantiates $q$ through a
    concrete case, mechanism, or example.
\end{itemize}

Figure~\ref{fig:pipeline} (left) gives one example per move and a non-match.
\broadrel{} and \specrel{} represent
opposite movements along an abstraction hierarchy, while \addrrel{} represents a functional response to a stated research need. In our experiments, we train a separate retriever $s_r$ for each operation, while our task formulation and data collection readily support training a relation-conditioned scoring function 
$s(q,c \mid r)$; see Section~\ref{sec:methods}.
Throughout, we use ``inspiration'' operationally to denote a statement that validly instantiates the requested ideation move. As in recent work \cite{garikaparthi-etal-2025-mir}, \system{} focuses its evaluation on this retrieval capability, not the downstream effect of retrieved statements on idea generation or their novelty to a particular researcher.

\section{Benchmark Construction}
\label{sec:benchmark_construction}
We mine pairs of a query and its gold sentence, the sentence that answers it, from consecutive sentences linked by a discourse marker. Authors often open a sentence with a marker that signals how the sentence relates to the preceding one, such as \emph{To mitigate these risks}, \emph{Even more specifically}, or \emph{From a broader perspective}. Consider the passage \emph{``Our model is prone to overfitting. To address this issue, we apply dropout.''} The first sentence is the query $q$, \emph{To address this issue} is the marker $\ell$, and the rest of the second sentence, \emph{we apply dropout}, is the gold sentence $g$. Markers thus provide high-precision, author-generated distant supervision: the author presents the second sentence as a response to the problem stated in the first. Figure~\ref{fig:pipeline} (right) summarizes the construction pipeline.

We use the full-text computer science papers of the Semantic Scholar corpus \citep{Kinney2023TheSS},\footnote{Snapshot of May 5, 2026.} after sentence segmentation and cleaning (Appendix~\ref{app:data-details}). For each move $r$, we build a lexicon $L_r$ of validated markers (\S\ref{sec:discourse-markers}). The three lexicons are disjoint, so each marker determines exactly one move. We scan all pairs of consecutive sentences $(u_i, u_{i+1})$ in each paper $P$. Whenever $u_{i+1} = \ell \oplus g$ for a marker $\ell \in L_r$, where $\oplus$ denotes concatenation, we extract the tuple $(u_i, \ell, g)$:
\begin{equation}
\mathcal{E}_r(P) = \{ (u_i, \ell, g) \mid u_{i+1} = \ell \oplus g,\ \ell \in L_r \}
\label{eq:extraction}
\end{equation}
We keep the marker $\ell$ only as metadata, so that a retriever cannot recognize the gold sentences of a move by their opening phrase.

\subsection{Ideation-Move Marker Lexicons}
\label{sec:discourse-markers}
We consider a phrase a valid marker of move $r$ if it meets two conditions: it signals $r$, and it refers back to the preceding sentence. For example, \emph{To mitigate this limitation} is an \addrrel{} marker, because it announces a remedy and \emph{this limitation} points to the problem stated in the preceding sentence. The phrase \emph{Therefore}, which prior work used to build NLI pairs \citep{sadat-caragea-2022-scinli}, fails the first condition, because it signals none of our three moves. The phrase \emph{In general} signals generalization but fails the second condition, because it does not refer to a specific preceding sentence; the \broadrel{} marker \emph{More generally}, in contrast, compares what follows with what was just said.

We collected candidate markers from the four sources below (details in Appendix~\ref{app:marker-lexicon-generation}, Table~\ref{tab:discourse-markers-full}).

\textbf{Manual curation.} Markers at the start of a sentence usually end with a comma, which separates them from the content of the sentence. For every sentence in the corpus, we therefore took the span before the first comma, if it had at most seven words, and counted how often each span occurred. Two NLP experts inspected the 400 most frequent spans. For each span, they read 15 randomly sampled sentence pairs whose second sentence begins with it. A span became a marker of a move only if both experts found that, in all 15 pairs, it clearly signaled that move and referred to the preceding sentence. This yielded 84 markers for \addrrel{}, 13 for \broadrel{}, and 72 for \specrel{}.

\textbf{Marker extensions.} For a few frequent prefixes, such as \emph{To address}, we also accepted short extensions up to a comma, such as \emph{To address this potential bias} (Appendix~\ref{app:marker-lexicon-generation}), and counted each prefix with its extensions as one marker.

\textbf{LLM generation.} Three LLMs from different model families proposed further \addrrel{} and \broadrel{} markers, starting from the manually curated ones. Using several families keeps the lexicon from inheriting one model's phrasing preferences. An LLM then labeled each generated phrase as \emph{strong-true}, \emph{weak-true}, or \emph{false}, according to whether the phrase signals the move in almost all contexts, only in some contexts, or in none. We kept only the 3,779 strong-true phrases. An NLP expert screened a random sample of 1,500 of them and confirmed that each signals the target move and plausibly refers to the preceding sentence.

\textbf{Pattern expansion.} Following \citet{roller-etal-2018-hearst}, who extracted hypernym--hyponym pairs with Hearst-style templates \citep{hearst1992automatic}, we wrote templates for \addrrel{} and \broadrel{} and filled their slots from curated term lists. For example, the \addrrel{} template \emph{to [response term] [demonstrative] [problem term]} produces phrases such as \emph{to mitigate these risks}. GPT-5.4 then removed ungrammatical phrases. This stage yielded 397 markers, 297 of them new.

Finally, we matched the three lexicons, which together contain 4,252 markers, against all 366.6M pairs of consecutive sentences in the corpus. Only 809 markers match at least one pair, and together they yield about 3M query--gold pairs from 1.1M papers. Two NLP experts independently vetted these 809 markers and agreed that every one was valid and clear-cut. We did not expand the \specrel{} lexicon with LLMs or templates, because its curated and extended markers alone yield 2.8M pairs.

\subsection{Temporal Split and Shared Candidate Pool}
\label{sec:split}
After de-duplication, \system{} contains 3,017,476 query--gold pairs: 2,779,177 for \specrel{}, 222,707 for \addrrel{}, and 15,592 for \broadrel{}. We split them by the publication date of their paper. Training pairs come from papers published between 2015 and September 2025, validation pairs from October to December 2025, and test pairs from January to May 2026. This split enforces a strict approach against contamination: no test query or test candidate could have been seen during training, and all test papers appeared after the public release of every model we evaluate.

Within each split, all three moves share one candidate pool, built only from that split's papers. Besides the gold sentences of the three moves, the pool contains \emph{distractors}: sentences that begin with a phrase we rejected as a marker, such as \emph{On the other hand} (contrast) or \emph{In general}. Most distractors originally began with a phrase of contrast, similarity, continuation, or inference. As with gold sentences, we remove the opening phrase, so that a retriever cannot tell distractors from gold sentences by their first words. No distractor is the gold sentence of any query. The training, validation, and test pools contain 13.8M, 361K, and 404K sentences. With a separate pool per move, a retriever would never need to tell the moves apart. In the shared pool, it must also reject sentences that are on topic but realize a different move or none. Appendix~\ref{app:split-full} gives full statistics (Table~\ref{tab:temporal-split-full}) and explains these choices. In pilot experiments, non-temporal splits gave substantially higher scores, consistent with the models having seen the evaluation texts during pre-training. Pools padded with easy-to-reject random sentences instead of distractors also inflated scores.

\subsection{Silver Test Set}
\label{sec:silver_set_construction}
Mined pairs are noisy, because a marker can be present even when the query cannot be understood on its own, or when the gold sentence realizes the move poorly. We therefore build a \emph{silver} test set, which keeps only the pairs accepted by an LLM judge calibrated against human labels. This is our primary evaluation set, as opposed to the unfiltered test set, which contains all mined test pairs. We sampled 10K \specrel{} test pairs, stratified by marker so that frequent markers do not dominate the sample, and took all 14,020 \addrrel{} and 1,410 \broadrel{} test pairs. GPT-5.4 judged each pair with two prompts per move, seeing the query, the marker, and the gold sentence, and we kept a pair only if both prompts accepted it. The prompts require that the query and the gold sentence each be well formed and understandable on their own, and that the gold sentence realize the move in the right direction. For \addrrel{}, for instance, the query must state a problem, and the gold sentence must directly address that problem. For \broadrel{} and \specrel{}, the two sentences must share their main topic, and a pair is dropped if the gold sentence is at the same level of abstraction as the query or if the direction is reversed. For each move, we selected the two prompts from four to six prompt variants by their agreement with human labels on 100 pairs of that move. On 60 held-out pairs per move labeled by five annotators, the selected prompts agree with an individual annotator at a mean $F_1$ of .87--.91 (\specrel{}), .73--.76 (\addrrel{}), and .75--.81 (\broadrel{}). The annotators agree with one another at .90, .87, and .82 (Appendix~\ref{app:judge-details}, Table~\ref{tab:prompt-agreement}). The silver set contains 17,579 queries: 7,327 for \specrel{}, 9,668 for \addrrel{}, and 584 for \broadrel{}. Rejected pairs are not used as queries, but their gold sentences stay in the candidate pool, and silver queries are ranked against the full test pool of 404,371 sentences, excluding the query's own sentence.

\section{Experiments}
\label{sec:methods}

\begin{table*}[t]
  \centering
  \small
  \setlength{\tabcolsep}{3pt}
  \renewcommand{\arraystretch}{1.1}
  \resizebox{\textwidth}{!}{
  \begin{tabular}{ll|cccc|cccc|cccc}
    \hline
    & & \multicolumn{4}{|c|}{\textbf{\specrel}} & \multicolumn{4}{c|}{\textbf{\addrrel}} & \multicolumn{4}{c}{\textbf{\broadrel}} \\
    \cline{3-6}\cline{7-10}\cline{11-14}
    \textbf{Model} & \textbf{Setup} & R@1 & R@10 & M@10 & M@100 & R@1 & R@10 & M@10 & M@100 & R@1 & R@10 & M@10 & M@100 \\
    \hline
    BM25 & Unigram & 13.0 & 27.3 & 17.6 & 18.1 & 5.3 & 15.2 & 8.3 & 8.7 & 7.9 & 17.6 & 10.9 & 11.3 \\
    \hline
    \multirow{3}{*}{\shortstack{all-mpnet-base-v2}}
      & Baseline         & 12.3 & 26.6 & 16.9 & 17.5 & 6.1  & 16.5 & 9.2  & 9.7  & 11.6 & 24.3 & 15.6 & 16.1 \\
      & Tuned            & 28.5 & 54.3 & 37.0 & 37.8 & 12.8 & 31.7 & 18.7 & 19.4 & 16.3 & 32.5 & 21.8 & 22.4 \\
      & Tuned + rerank   & 45.3 & 66.0 & 52.4 & --   & 28.5 & 45.9 & 34.2 & --   & 23.8 & 37.7 & 27.8 & --   \\
    \hline
    \multirow{3}{*}{\shortstack{ModernBERT-embed-large}}
      & Baseline         & 20.5 & 39.1 & 26.7 & 27.3 & 6.7  & 17.8 & 10.2 & 10.6 & 13.4 & 26.0 & 17.3 & 17.9 \\
      & Tuned            & \textbf{37.2} & \textbf{65.5} & \textbf{46.7} & \textbf{47.4} & \textbf{17.2} & \textbf{40.0} & \textbf{24.5} & \textbf{25.3} & \textbf{20.9} & \textbf{41.4} & \textbf{27.3} & \textbf{27.8} \\
      & Tuned + rerank   & \underline{47.5} & \underline{69.9} & \underline{55.1} & --   & \underline{30.8} & \underline{52.1} & \underline{37.7} & --   & 22.3 & \underline{40.9} & 28.2 & --   \\
    \hline
    \multirow{3}{*}{\shortstack{Stella-en-1.5B-v5}}
      & Baseline         & 20.0 & 39.8 & 26.4 & 27.0 & 7.2  & 20.1 & 11.0 & 11.6 & 13.7 & 29.6 & 18.6 & 19.2 \\
      & Tuned            & 31.8 & 60.0 & 41.0 & 41.8 & 13.8 & 34.4 & 20.3 & 21.0 & 15.8 & 31.8 & 20.9 & 21.7 \\   
      & Tuned + rerank   & 45.6 & 68.0 & 53.4 & --   & 28.5 & 47.1 & 34.5 & --   & \underline{24.0} & \underline{40.9} & \underline{29.5} & --   \\
    \hline
    EmbeddingGemma-300m       & Baseline & 22.1 & 42.3 & 28.8 & 29.4 & 4.9 & 14.6 & 7.8  & 8.3  & 14.0 & 28.9 & 18.5 & 19.1 \\
    Qwen3-Embedding-0.6B      & Baseline & 17.6 & 37.0 & 24.0 & 24.5 & 6.8 & 18.9 & 10.4 & 10.9 & 12.0 & 27.1 & 16.5 & 17.1 \\
    ReasonIR-8B               & Baseline & 22.6 & 44.3 & 29.6 & 30.2 & 8.0 & 21.7 & 12.1 & 12.7 & 13.0 & 30.8 & 18.3 & 18.9 \\
    \hline
    answerai-colbert-small-v1 & Baseline & 15.6 & 30.7 & 20.5 & 21.0 & 4.2 & 11.0 & 6.3  & 6.6  & 7.4  & 15.8 & 10.0 & 10.3 \\
    GTE-ModernColBERT-v1      & Baseline & 17.4 & 34.3 & 22.7 & 23.2 & 5.5 & 14.9 & 8.4  & 8.8  & 7.5  & 20.9 & 11.9 & 12.4 \\
    Reason-ModernColBERT      & Baseline & 17.2 & 34.0 & 22.6 & 23.1 & 5.9 & 16.7 & 9.2  & 9.7  & 9.1  & 22.1 & 13.6 & 14.1 \\
    \hline
  \end{tabular}
  }
  \caption{\label{tab:results_silver}
    Results on the silver test set (\%); every query is ranked against the full test pool. R@$k$ = Recall@$k$, M@$k$ = MRR@$k$. \emph{Baseline}: pre-trained model; \emph{Tuned}: fine-tuned on the move; \emph{Tuned + rerank}: GPT-5-mini reranks a fine-tuned top 100 and returns the 10 best. ModernBERT-embed-large and Stella-en-1.5B-v5 use their recommended prefixes, all-mpnet-base-v2 a generic one (\S\ref{sec:retrievers}). Bold: best without reranking. Underline: best reranked. Table~\ref{tab:results_silver_full} (Appendix~\ref{app:full-results}) adds the generic-prefix setups, BM25 n-grams, reranked Baselines, and move instructions.
  }
\end{table*}

\subsection{Retrievers}
\label{sec:retrievers}
\textbf{One retriever per move.} We train one retriever per move to rank the shared candidate pool. The move determines which type of candidate to look for, but not which candidate fits the query. The \addrrel{} retriever, for example, must still separate candidates that address the query's core problem from topically related ones that do not. We use separate retrievers to isolate the learnability of each move and to represent a plausible modular system in which a user selects a specialized retrieval expert. A unified model trained with $r$ as an instruction is a complementary setting supported by the benchmark and left for future work.

\textbf{Models.} We fine-tune three dense retrievers: all-mpnet-base-v2 \citep{all-mpnet-base-v2}, ModernBERT-embed-large \citep{ModernBERT-embed-large}, and Stella-en-1.5B-v5 \citep{Zhang2024JasperAS}. The last, which we call Stella, performs best on MIR \citep{garikaparthi-etal-2025-mir}. All three were pre-trained on large amounts of scientific text. We fine-tune each model separately for each move with a contrastive loss, in which the negatives for a query are the gold sentences of the other queries in the batch. All three models are trained with a generic input prefix, and ModernBERT-embed-large and Stella also with the prefix recommended by their developers. These five combinations of model and prefix, trained for each move, give 15 fine-tuned retrievers, which took 1,800 GPU hours to train. For each, we tried three to five hyperparameter configurations and selected the run and checkpoint by validation MRR@10; we used the test set only once, for the final results (Appendix~\ref{app:training-details}). We compare each fine-tuned retriever with its \emph{baseline}, the same model before fine-tuning, and with BM25. We also evaluate six other pre-trained retrievers without fine-tuning, all with the generic prefix. Three are dense retrievers: EmbeddingGemma-300m \citep{vera2025embeddinggemmapowerfullightweighttext}, Qwen3-Embedding-0.6B \citep{zhang2025qwen3embeddingadvancingtext}, and ReasonIR-8B \citep{shao2025reasonir}, trained for reasoning-intensive retrieval. Three are late-interaction retrievers: answerai-colbert-small-v1 \citep{clavie2024jacolbertv2}, GTE-ModernColBERT-v1 \citep{GTE-ModernColBERT}, and Reason-ModernColBERT \citep{Reason-ModernColBERT}. Surprisingly, a move instruction changes the MRR@10 of three instruction-following baselines, Qwen3-Embedding-0.6B, ReasonIR-8B, and Stella, by only $-2.9$ to $+2.6$ points (Appendix~\ref{app:training-details}). Finally, GPT-5-mini reranks the top 100 candidates of the three fine-tuned retrievers and their baselines.

\subsection{Results on the Silver Test Set}
\label{sec:results}
Table~\ref{tab:results_silver} reports results on the silver test set; MRR@10 is our primary metric. We omit Precision@$k$ and nDCG@$k$: with a single gold sentence per query, the first equals Recall@$k$ divided by $k$, and the second depends only on the rank of the gold sentence. Without fine-tuning, the best retriever for each move in Table~\ref{tab:results_silver} exceeds BM25 by 12.0 (\specrel{}), 3.8 (\addrrel{}), and 7.7 (\broadrel{}) MRR@10 points, and some retrievers score below BM25, whose bigram and trigram variants score lower still (Appendix~\ref{app:full-results}). Fine-tuning ModernBERT-embed-large adds 20.0, 14.3, and 10.0 points over its baseline on \specrel{}, \addrrel{}, and \broadrel{}, respectively, which multiplies its MRR@10 by 1.7, 2.4, and 1.6. After fine-tuning, it is the strongest retriever for every move, and its generic and recommended prefixes give similar results, within 2.7 points on every move and metric (Appendix~\ref{app:full-results}). Even so, it ranks the gold sentence in its top 10 for only 40.0\% of \addrrel{} queries and 41.4\% of \broadrel{} queries. These results show the task is far from solved. On the unfiltered test set, fine-tuning improves every model too, raising the MRR@10 of ModernBERT-embed-large from 21.7, 9.7, and 15.5 to 39.1, 21.5, and 22.2, and it is again the strongest retriever on every move (Appendix~\ref{app:full-results}).

\textbf{LLM reranking.} Reranking the top 100 candidates of the fine-tuned ModernBERT-embed-large raises MRR@10 to 55.1, 37.7, and 28.2 on \specrel{}, \addrrel{}, and \broadrel{}, respectively (Table~\ref{tab:results_silver}). Reranking helps all three fine-tuned retrievers, adding 8.4 to 15.5 MRR@10 points on \specrel{} and \addrrel{}, and on \specrel{} it narrows the gap between the best and worst of them from 9.7 to 2.7 points. On \specrel{} and \addrrel{}, fine-tuning and reranking are complementary for all three retrievers. For ModernBERT-embed-large, together they add 28.4 and 27.5 points over the baseline, well above either alone (at most 20.0 and 14.3). The reranker can only promote a gold sentence from the retriever's top 100, so first-stage recall bounds its gains. The fine-tuned ModernBERT-embed-large places the gold sentence there for 80.8\% of \specrel{}, 61.3\% of \addrrel{}, and 57.2\% of \broadrel{} queries, and for these queries the reranker ranks it in its top 10 in 86.6\%, 85.1\%, and 71.6\% of cases, respectively.

\textbf{Oracle move filtering.} A fine-tuned retriever could succeed by a shortcut: it could first recognize sentences written in the style of the requested move (for example, sentences that describe a method) and then choose the one closest in topic to the query. To test for this shortcut, we use the move labels as an oracle and restrict the test pool to the gold sentences of the requested move, so that every candidate has the style of that move. In this restricted pool (94,079, 14,020, and 1,410 candidates), fine-tuning ModernBERT-embed-large still adds 22.1, 19.4, and 14.8 MRR@10 points on \specrel{}, \addrrel{}, and \broadrel{}, respectively, and 26.6, 20.1, and 15.8 points when the pool holds only the gold sentences of the silver queries (Appendix~\ref{app:oracle-filtering}). The style of a move may still serve as a coarse cue, since it is part of the supervision, but it cannot alone explain the gains. Nor can lexical matching: in all 24 comparisons, which cover four cutoffs $k$, three moves, and the unfiltered and silver sets, the fine-tuned retriever's top-$k$ candidates share fewer words with the query than the baseline's do. On \specrel{}, its top-10 candidates share even fewer than the gold sentence does (Appendix~\ref{app:analysis-details}). Nor can a next-sentence skill common to all moves: on \specrel{} and \addrrel{} queries, each retriever scores best when fine-tuned on that move (Appendix~\ref{app:cross-move}).

\begin{table}[t]
\centering\small
\setlength{\tabcolsep}{3pt}
\resizebox{\columnwidth}{!}{%
\begin{tabular}{llrrrrr}
\toprule
Move & Setup & HR & P & MRR & Pref. & Tie \\
\midrule
\multirow{2}{*}{\specrel{} (39)} & Baseline & 50.0 & 12.7 & 37.4 & 3 & \multirow{2}{*}{10} \\
 & Tuned & \textbf{87.2} & \textbf{20.6} & \textbf{63.9} & \textbf{29} & \\
\midrule
\multirow{2}{*}{\addrrel{} (40)} & Baseline & 42.5 & 7.0 & 21.7 & 4 & \multirow{2}{*}{7} \\
 & Tuned & \textbf{85.0} & \textbf{19.5} & \textbf{42.3} & \textbf{30} & \\
\midrule
\multirow{2}{*}{\broadrel{} (46)} & Baseline & 50.0 & 9.3 & 36.8 & 8 & \multirow{2}{*}{18} \\
 & Tuned & \textbf{72.8} & \textbf{16.0} & \textbf{49.5} & \textbf{22} & \\
\midrule
\multirow{2}{*}{All (125)} & Baseline & 47.6 & 9.6 & 32.1 & 15 & \multirow{2}{*}{35} \\
 & Tuned & \textbf{81.2} & \textbf{18.6} & \textbf{51.7} & \textbf{81} & \\
\bottomrule
\end{tabular}%
}
\caption{Human inspiration evaluation of the top 10 candidates of ModernBERT-embed-large with the \texttt{search\_query} prefix, before (\emph{Baseline}) and after (\emph{Tuned}) fine-tuning (\%, all metrics at rank 10, mean over queries). A candidate counts as relevant if the annotator marked it useful. The number of queries is in parentheses. HR = hit rate, P = precision. All differences in HR, P, and MRR between Baseline and Tuned are significant ($p<.01$, paired permutation test). Pref.: number of annotations that preferred the list; Tie: number of annotations that preferred neither. Bold: the better of Baseline and Tuned for each move.}
\label{tab:human_eval}
\end{table}

\subsection{Human Inspiration Evaluation}
\label{sec:human_inspiration}
The silver metrics credit a retriever only for the rank of the gold sentence. Other sentences in the pool may inspire a researcher as well, but the silver metrics count them as wrong and can therefore underestimate retrieval quality, as \citet{garikaparthi-etal-2025-mir} note for MIR. Nor can these metrics show whether the retriever finds inspirations beyond the sentence that follows the query in its own paper. We therefore asked researchers to judge the retrieved candidates as inspirations. We evaluate ModernBERT-embed-large, the strongest retriever without reranking (Table~\ref{tab:results_silver}). Five computer science graduate students, each with at least one publication, labeled 125 silver queries (39 \specrel{}, 40 \addrrel{}, and 46 \broadrel{}), assigned to them by subfield to simulate a researcher who seeks inspiration in their own area. Annotators were blind to condition: for each query, they saw the top 10 candidates of ModernBERT-embed-large before and after fine-tuning as two lists side by side, labeled List A and List B, in shuffled order. A candidate is useful if it would help the annotator address the query (\addrrel{}), broaden it (\broadrel{}), or find a concrete example of it (\specrel{}). The annotator marked each candidate as useful or not and chose the more useful list, allowing ties. Annotators were trained with guidelines, a video, and an example labeled with the first author. Each annotator labeled 18 to 33 queries, and annotation took a little over 30 hours in total (Appendix~\ref{app:human-eval-top10}).

\textbf{Results.} HR@10 is the fraction of queries with a useful candidate in the top 10, and MRR@10 is the mean reciprocal rank of the first useful candidate. Fine-tuning raises HR@10 and MRR@10 on every move ($p<.01$, paired permutation test; Table~\ref{tab:human_eval}). Over all 125 queries, HR@10 rises from 47.6 to 81.2 and MRR@10 from 32.1 to 51.7. Annotators preferred the fine-tuned list more often than the baseline list on every move: 29 vs.\ 3 times on \specrel{}, 30 vs.\ 4 on \addrrel{}, and 22 vs.\ 8 on \broadrel{} (73--91\% of non-tie choices).

\textbf{Inspirations beyond the gold sentence.} Annotators marked the gold sentence useful in 96.5\% of the cases in which it appeared in the fine-tuned top 10, which supports the silver labels. Most useful candidates, however, are not the gold sentence. For 64.8\% of the queries, the fine-tuned top 10 contains a useful candidate other than the gold sentence (62.8\% on \specrel{}, 70.0\% on \addrrel{}, and 62.0\% on \broadrel{}), against 42.0\% for the baseline, which is lower on every move (Appendix~\ref{app:human-beyond-gold}, Table~\ref{tab:human_beyond_gold}). Of these candidates, 91.7\% come from a paper other than the query's (81.4\% for the baseline), close to the 95.4\% of all non-gold candidates in the top 10 (Table~\ref{tab:crosspaper-app}). The fine-tuned retriever's 180 such candidates come from 176 distinct papers. Over the same 125 queries, the fine-tuned retriever reaches MRR@10 of 51.7 when every useful candidate counts, against 27.1 when only the gold sentence counts; per move and retriever, this ratio is 1.5 to 3.1. The fine-tuned retriever thus finds inspirations in other papers, beyond the sentence that follows the query in its own paper. The candidate-level human judgments will be released as reusable relevance labels.

\section{Conclusion}
\label{sec:conclusion}
We introduced \system{}, a benchmark for scientific inspiration retrieval where relevance depends on ideation roles. We establish ideation operation-conditioned retrieval and provide scalable resources for exploring it. 
\system{} focuses on three operations: \addrrel{} retrieves responses to problems; \broadrel{} retrieves more general formulations; and \specrel{} retrieves more concrete instances. \system{} contains over 3 million query--positive pairs from full-text papers, using vetted discourse markers. Human-calibrated LLM judgments provide a higher-quality silver test set. 

Our experiments show that retrievers do not capture these ideation moves, while move-specific fine-tuning improves all models and LLM reranking adds further gains. Tuned models frequently retrieve useful inspirations other than the mined positive, especially for \addrrel{}. Overall performance leaves substantial room for improvement.

Although we instantiate \system{} for three ideation moves in CS, our methodology can be extended to other relations and other domains, offering a general recipe for operation-conditioned retrieval supervision. Future work could extend the task beyond adjacent sentences and CS, collect extensive (costly) human annotations, and integrate these retrieval operations into models and systems for scientific reasoning and idea development.

\clearpage
\section*{Limitations}

Marker coverage is uneven across operations. We evaluate three fundamental ideation moves, but future work could explore more, such as analogy, which builds on the abstraction that \broadrel{} captures, and contrast, which is frequent in the text.

The benchmark evaluates whether retrieved statements realize a requested ideation move; it does not determine whether an inspiration is novel---this is orthogonal to our work. We evaluate retrieval, not its downstream effect on ideation.

Because the query sets are mined separately for each relation as naturally occurring instances in scientific papers, the present experiments evaluate specialization within each operation-specific query distribution and not a scenario where the same query is issued under multiple operations---this would require expensive annotation which is outside the scope of this paper.

Pairs are mined only where authors signal the move explicitly with a marker. The single preceding sentence misses antecedents that span a longer passage; future work could apply (costly) LLM expansion of the query/gold contexts. Markers for the reverse, solution-then-problem order (e.g., \emph{``This addresses,''}) fit the same recipe but are far rarer. Each released pair carries the Semantic Scholar corpus ID and publication date of its paper, from which further metadata can be easily extracted.

The mined gold always comes from the query's own paper. Retrieval runs over a shared pool of 404K sentences, and most useful candidates come from other papers (\S\ref{sec:human_inspiration}); a test set built from citation links between papers is future work.

The candidate pool does not contain every sentence of the corpus. Evaluating retrieval over every sentence of the corpus, which would test coverage at deployment scale, is left to future work.

\clearpage
\bibliography{custom}
\clearpage

\appendix
\raggedbottom

\section{Marker Lexicons: Full Construction}
\label{app:marker-lexicon-generation}

\paragraph{Annotators} Two NLP experts vetted the markers: a graduate student in computer science with experience in text annotation, and a university professor. A separate group of annotators labeled the sentence pairs that we used to calibrate the LLM judge (Appendix~\ref{app:judge-details}).

\paragraph{Manual curation}
In scientific writing, markers of our three moves usually open a sentence and end with a comma, which separates them from the content of the sentence. For every sentence in the corpus, we therefore took the span before the first comma, if it had at most seven words, and counted how often each span occurred. We treated the 400 most frequent spans as candidate markers. For each span, two NLP experts read 15 randomly sampled sentence pairs whose second sentence starts with it. A span became a marker of a move only if both experts found that, in all 15 pairs, it clearly signaled that move and referred to the preceding sentence (the two conditions of \S\ref{sec:discourse-markers}). In total, 84 spans became \addrrel{} markers, 13 became \broadrel{} markers, and 72 became \specrel{} markers.

\paragraph{Marker extensions}
We also selected by hand a few frequent prefixes, which we call \emph{extendable connectives}, and automatically added their extensions as markers. For the prefix \emph{To address}, for example, we collected every extension of at most five words that ends at a comma, so that \emph{To address this potential bias}, \emph{To address this limitation}, and similar phrases were all added as markers. Table~\ref{tab:discourse-markers-full} counts each extendable connective as a single marker, however many extensions it has.

\paragraph{LLM generation}
Three LLMs from different model families, namely Claude Opus~4.6 \citep{Anthropic2026Claude}, Gemini~3 Flash \citep{Google2026Gemini}, and GPT-5.4 \citep{OpenAI2026GPT54}, proposed new \addrrel{} and \broadrel{} markers over several rounds, starting from the manually curated markers. We used three model families so that the lexicon does not inherit the phrasing preferences of a single model. Each LLM received the definitions of the moves, seed phrases built from the Hearst-style templates described under \emph{Pattern expansion} below, and the subcategories of each move. The subcategories of \addrrel{} were types of response to a problem, such as \emph{alleviate}, \emph{avoid or eliminate}, \emph{bypass}, \emph{mitigate}, and \emph{solve}. An LLM then labeled each of the 4,256 generated candidates as \emph{strong-true} if it signals the move in almost all contexts, as \emph{weak-true} if it signals the move only in some contexts, or as \emph{false} if it does not signal the move. The LLM assigned these labels to 3,779, 270, and 207 candidates, respectively. We kept only the 3,779 strong-true candidates, which make up 89\% of the final lexicon of 4,252 markers. An NLP expert then screened a random sample of 1,500 of them and confirmed that each signals the target move and plausibly refers to the preceding sentence. Appendix~\ref{sec:discourse_markers_classifier} gives the labeling prompts for \addrrel{} and \broadrel{}.

\paragraph{Pattern expansion}
\citet{hearst1992automatic} observed that some lexico-syntactic patterns reliably signal an is-a relation in running text. For example, in \emph{``[NP$_1$] such as [NP$_2$]''}, NP$_1$ is almost always a hypernym of NP$_2$, so such patterns can mine hypernym--hyponym pairs from a corpus without supervision \citep{roller-etal-2018-hearst}. We apply the same idea to discourse relations. Each template consists of fixed text and typed slots. For example, the \addrrel{} template \texttt{to [solT] [dem-pron] [probT]} has slots for a response term, a demonstrative, and a problem term. We filled each slot from a list of terms of the slot's type. We then used NLTK, spaCy, and PyInflect to expand each filled template into its morphological variants, so that one template yields every inflected form of its verbs and nouns. GPT-5.4 then removed ungrammatical candidates. Table~\ref{tab:expansion-patterns} lists example patterns.

For \addrrel{}, templates combining 116 problem terms, 81 response terms, and 18 modifiers generated about 6K candidates. For \broadrel{}, two generalization verbs (\emph{generalize}, \emph{broaden}) combined with ten nouns for what is generalized (\emph{observation, result, finding, analysis, case, phenomenon, insight, experiment, example, problem}) produced 810 candidates. This stage yielded 397 markers. Because 100 of them had already been produced by the earlier sources, pattern expansion added 297 new markers, 257 for \addrrel{} and 40 for \broadrel{}.

\begin{table}[t]
\centering
\small
\setlength{\tabcolsep}{3pt}
\renewcommand{\arraystretch}{1.15}
\begin{tabular}{@{}
  >{\raggedright\arraybackslash}p{0.55\columnwidth}
  >{\raggedright\arraybackslash}p{0.39\columnwidth}@{}}
\toprule
\textbf{Pattern} & \textbf{Example} \\
\midrule
\multicolumn{2}{@{}l}{\addrrel{}} \\
(in order) to [response] this ([problem]) & in order to mitigate this limitation \\
{}[response]-\emph{ing} this ([problem]) & addressing this challenge \\
a ([modifier]) [response]-\emph{noun} for this ([problem]) & a possible solution for this issue \\
\midrule
\multicolumn{2}{@{}l}{\broadrel{}} \\
(to) [generalize]-\emph{ing} this & generalizing this \\
to [generalize] this [case] & to generalize these observations \\
{}[broaden cue] this & zooming out from this \\
{}[report verb] this \{more generally $|$ more broadly\} & suggests this more generally \\
this [report verb] \{in general $|$ a more general\} & this indicates a more general \\
\bottomrule
\end{tabular}
\caption{Examples of expansion patterns for \addrrel{} and \broadrel{}, the two moves to which we applied pattern expansion. Square brackets mark slots filled from curated word lists, parentheses mark optional elements, and braces mark alternatives, of which one is chosen. \emph{-ing} and \emph{-noun} mark the gerund and the noun form of the slot filler (e.g., \emph{addressing}, \emph{solution}). \emph{this} stands for any demonstrative (\emph{this}, \emph{that}, \emph{these}, \emph{those}) and agrees in number with the following noun. Patterns are stored in lemmatized form and expanded by inflection; we list them in inflected form for readability.}
\label{tab:expansion-patterns}
\end{table}

\paragraph{Lexicon size and corpus matching}
Table~\ref{tab:discourse-markers-full} breaks the lexicon down by source and move. We matched the full lexicon $L = L_{\addrrel} \cup L_{\broadrel} \cup L_{\specrel}$ against all $\sim$366.6M pairs of consecutive sentences. For each pair, we checked whether the second sentence begins with a marker $\ell \in L$. If it does, the first sentence becomes the query $q$, the rest of the second sentence becomes the gold sentence $g$, and the pair is assigned to the unique move $r$ with $\ell \in L_r$ (the lexicons are disjoint by construction). Of the 4,252 markers, 809 occur at least once (the \emph{Data} columns of Table~\ref{tab:discourse-markers-full}), and together they yield $\sim$3M pairs from 1.1M papers. Two NLP experts independently vetted these 809 markers, and both judged all of them to be valid and clear-cut markers of their move. Because manual curation and extendable connectives alone gave 2.8M \specrel{} pairs, we did not apply LLM generation or pattern expansion to \specrel{}.

\begin{table}[t]
\centering
\small
\setlength{\tabcolsep}{3pt}
\resizebox{\columnwidth}{!}{
\begin{tabular}{@{}lrrrrrrrr@{}}
\toprule
&\multicolumn{2}{c}{\addrrel{}}
&\multicolumn{2}{c}{\broadrel{}}
&\multicolumn{2}{c}{\specrel{}}
&\multicolumn{2}{c}{Total}\\
\cmidrule(lr){2-3}\cmidrule(lr){4-5}\cmidrule(lr){6-7}\cmidrule(lr){8-9}
Source&Lexicon&Data&Lexicon&Data&Lexicon&Data&Lexicon&Data\\
\midrule
Manual&84&49&13&7&72&52&169&108\\
Extendable connective&5&4&0&0&2&2&7&6\\
LLM-generated&3516&473&263&25&0&0&3779&498\\
Pattern expansion&257&191&40&6&0&0&297&197\\
\midrule
Total&3862&717&316&38&74&54&4252&809\\
\bottomrule
\end{tabular}
}
\caption{Discourse markers by source and move. \emph{Lexicon}: markers in the lexicon. \emph{Data}: markers that occur in at least one extracted pair. \emph{Extendable connective}: a prefix, such as \emph{To address}, whose extensions of at most five words up to a comma were all added as markers; each prefix is counted as one marker.}
\label{tab:discourse-markers-full}
\end{table}

\paragraph{Distractor markers}
Distractor markers are sentence-opening phrases that we rejected as markers, so they belong to none of the three lexicons. The sentences they introduce resemble gold sentences in genre and position and are extracted in the same way, but they do not realize any of the three moves. Most distractor markers fall into two groups. The first group contains markers that do not reliably refer to the immediately preceding sentence, such as \emph{In general} and \emph{Therefore}. The phrase \emph{Therefore} also signals none of our moves (\S\ref{sec:discourse-markers}), and it often points back to an earlier paragraph. The second group contains markers of other relations that are very common in our corpus, such as contrast (\emph{In contrast}) and similarity (\emph{Similarly}, \emph{In other words}). An NLP expert checked 1,000 of the distractor markers and confirmed that none of them is a valid marker of any of the three moves. An LLM performed the same check on the remaining distractor markers.

\paragraph{Released lexicon}
We release the full lexicon: the markers of the three moves and the distractor markers, grouped into the four distractor categories of Table~\ref{tab:temporal-split-full}.

\section{Data Extraction and Temporal Split}
\label{app:data-details}

\subsection{Text Normalization and Cleaning}
\begin{figure*}[t]
    \centering
    \includegraphics[width=1\linewidth]{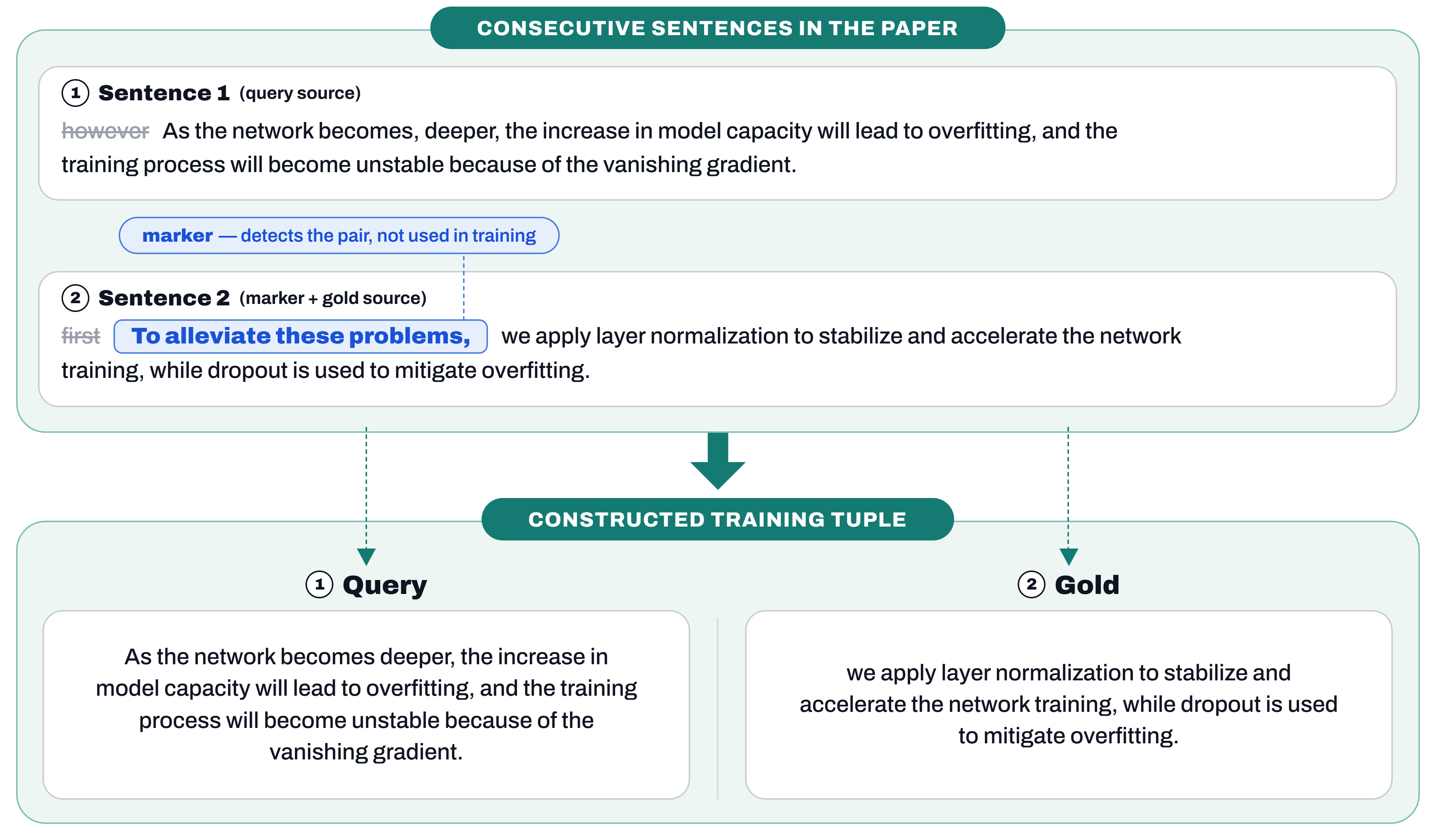}
    \caption{Example of extracting a query, a marker, and a gold sentence from a paper, including the removal of the marker and the cleaning steps.}
    \label{fig:s2orc-example}
\end{figure*}

We clean the extracted sentences with the following regular-expression heuristics:
\begin{itemize}[nosep, leftmargin=1.2em]
\item \textbf{Normalization:} we lowercase the text.
\item \textbf{Noise removal:} we remove malformed text, invalid characters, redundant whitespace, and stray list markers, such as bracketed numbers, leading ordinals, and prefixes like \texttt{1)}.
\item \textbf{Query prefix removal:} we remove a leading discourse marker, such as \emph{however}, \emph{first}, or \emph{in particular}, from the first sentence of a pair, which becomes the query.
\item \textbf{Gold-sentence prefix removal:} we remove ordinal and list prefixes, such as \emph{firstly} and \emph{second}, from gold sentences, so that each gold sentence starts with its content.
\item \textbf{Short-sentence removal:} we drop sentences shorter than 35 characters; we inspected all of them manually.
\end{itemize}
Each stored tuple also records metadata: the corpus ID (\texttt{corpusid}), the publication date, field-of-study labels, the section category, and the paragraph. Figure~\ref{fig:s2orc-example} shows an example of the extraction.

\paragraph{License} The Semantic Scholar corpus is distributed under the ODC-By 1.0 license, which allows redistribution of derived data with attribution. We release our data and annotations under the same license.

\subsection{Temporal Split and Shared Candidate Pool}
\label{app:split-full}

\paragraph{Split} We first de-duplicated the dataset, so that each $(q, g)$ pair appears at most once, and then partitioned the pairs by the publication date of their paper. The training set contains pairs from papers published from 2015 to September 30, 2025, the validation set from October 1 to December 31, 2025, and the test set from January 1 to May 5, 2026. Each split contains three disjoint sets of queries, one per move. Because the splits do not overlap in time, no test query or test candidate is seen during fine-tuning. The test papers were also published after the public release of every model we evaluate. Table~\ref{tab:temporal-split-full} gives the size of each partition.

\paragraph{Why a temporal split} The split tests whether retrievers trained on earlier literature generalize to newly published work. In pilot experiments, non-temporal splits, including a split by marker, gave substantially higher scores. These higher scores are consistent with the models having seen the evaluation texts, or close versions of them, during pre-training. Taking both the test queries and the test candidates from recent papers reduces the chance that the models saw the test texts during pre-training. The short gap between training and test papers also limits shifts in the distribution of computer science topics.

\paragraph{Split dates} Three considerations determined the split dates. An earlier training cutoff would place the evaluation data inside the likely pre-training periods of the models. Assigning all of 2025 to validation would make the validation set unnecessarily large, because computer science publishing grew rapidly in 2025. A shorter test period would leave too few \broadrel{} test pairs.

\paragraph{Shared candidate pool} Within each split, the three moves share a single candidate pool. The gold sentence of a query is the sentence that follows the query in its paper, with the marker removed. This gold sentence competes against every other sentence in the pool. A retriever therefore cannot assume that the pool contains only candidates of the requested move. For example, the pool that an \addrrel{} retriever searches also contains the gold sentences of \specrel{} and \broadrel{} queries. Besides the gold sentences of the three moves, each pool contains \emph{distractors}. A distractor is a sentence introduced by a \emph{distractor marker}, a phrase that we rejected as a marker during curation (Appendix~\ref{app:marker-lexicon-generation}). No distractor is the gold sentence of a query. The pool contains neither every sentence of the corpus nor only the gold sentences of the three moves. Gold sentences make up only about 20\% of the 13.8M training candidates. The remaining 80\% are distractors from more than one million papers, which follow markers of contrast, similarity, continuation, inference, and other common relations. The pool is therefore highly heterogeneous in topic, rhetorical function, and linguistic form, and it is substantially closer to broad scientific-sentence retrieval than to ranking within a small, move-specific collection. Distractors are extracted with the same procedure as gold sentences, and they often resemble gold sentences in topic, in style, and in the discourse cues that remain after the marker is removed. Because the pool is shared across moves, it also contains sentences that are on topic for a query but realize a different move. A retriever must therefore judge whether a candidate realizes the requested move for the query, not merely reject malformed, out-of-domain, or off-topic sentences. Random corpus sentences would mostly add such easy negatives. In pilot experiments, pools padded with random sentences instead of distractors gave inflated scores.

\begin{table*}[!t]
\centering
\small
\setlength{\tabcolsep}{4pt}
\resizebox{\textwidth}{!}{%
\begin{tabular}{lrrcrrcrrc}
\toprule
\multirow{2}{*}{Source}
&\multicolumn{3}{c}{Train}
&\multicolumn{3}{c}{Validation}
&\multicolumn{3}{c}{Test}\\
\cmidrule(lr){2-4}\cmidrule(lr){5-7}\cmidrule(lr){8-10}
&Queries&Sentences&Candidates&Queries&Sentences&Candidates&Queries&Sentences&Candidates\\
\midrule
\multicolumn{3}{l}{\emph{Target moves}}&\multirow{11}{*}{13,787,834}&\multicolumn{2}{c}{}&\multirow{11}{*}{361,172}&\multicolumn{2}{c}{}&\multirow{11}{*}{404,371}\\
\specrel{}&2,605,515&2,605,515&&79,583&79,583&&94,079&94,079&\\
\addrrel{}&195,605&195,605&&13,082&13,082&&14,020&14,020&\\
\broadrel{}&13,687&13,687&&495&495&&1,410&1,410&\\
\cmidrule(lr){1-3}\cmidrule(lr){5-6}\cmidrule(lr){8-9}
\multicolumn{3}{l}{\emph{Distractors}}&&\multicolumn{2}{c}{}&&\multicolumn{2}{c}{}&\\
Explicit contrast&--&1,135,357&&--&48,963&&--&63,086&\\
Implicit similarity&--&15,440&&--&426&&--&401&\\
Explicit similarity&--&567,983&&--&12,636&&--&12,518&\\
Common irrelevant&--&9,254,247&&--&205,987&&--&218,857&\\
\midrule
All&2,814,807&13,787,834&13,787,834&93,160&361,172&361,172&109,509&404,371&404,371\\
\bottomrule
\end{tabular}%
}
\caption{Full temporal split. Training uses papers published from 2015 to Sept.~30, 2025, validation from Oct.~1 to Dec.~31, 2025, and test from Jan.~1 to May~5, 2026. Each query has a single gold sentence. The Sentences column counts the sentences that each source contributes to the candidate pool: gold sentences for the three moves, and distractors for the four distractor categories. Distractors are never the gold sentence of a query and are never used as queries. The candidate pool of each split is shared by all moves and contains all of these sentences.}
\label{tab:temporal-split-full}
\end{table*}

\section{Training and Evaluation Setup}
\label{app:training-details}

\paragraph{Models and software}
We fine-tune all-mpnet-base-v2\footnote{\url{https://huggingface.co/sentence-transformers/all-mpnet-base-v2}}, ModernBERT-embed-large\footnote{\url{https://huggingface.co/lightonai/modernbert-embed-large}} \citep{modernbert,ModernBERT-embed-large}, and Stella-en-1.5B-v5\footnote{\url{https://huggingface.co/NovaSearch/stella_en_1.5B_v5}} \citep{Zhang2024JasperAS} with the \texttt{sentence-transformers} library \citep{reimers-2019-sentence-bert}. 

\paragraph{Loss}
We train with the multiple negatives ranking loss, a contrastive loss with in-batch negatives \citep{henderson-etal-2017-efficient,oord2019representationlearningcontrastivepredictive}:
\begin{equation}
\mathcal{L}
= -\frac{1}{B}\sum_{i=1}^{B}
\log
\frac{e^{s\cos(\mathbf{q}_i,\mathbf{g}_i)}}
{\sum_{j=1}^{B} e^{s\cos(\mathbf{q}_i,\mathbf{g}_j)}},
\end{equation}
where $\mathbf{q}_i$ and $\mathbf{g}_i$ are the embeddings of the $i$-th query and its gold sentence in a batch of $B$ pairs, $\cos$ is cosine similarity, and $s$ is a scale factor. We use the library default $s = 20$, which corresponds to a temperature of $\tau = 1/s = .05$.

\paragraph{Input prefixes}
All three models are trained with a \textbf{generic prefix}: \texttt{"query: "} before each query and \texttt{"document: "} before each candidate. ModernBERT-embed-large is also trained with its recommended prefixes, \texttt{search\_query:} for queries and \texttt{search\_document:} for candidates, and Stella with the query prefix recommended by its developers. Over the three moves, these five combinations of model and prefix give 15 fine-tuned retrievers.

\paragraph{Move instructions}
Qwen3-Embedding-0.6B, ReasonIR-8B, and Stella-en-1.5B-v5 are trained to follow instructions, so we also run them, without fine-tuning, with a move-specific instruction:
\begin{itemize}[nosep, leftmargin=1.2em]
    \item \addrrel{}: \emph{Given a scientific statement describing a research problem or limitation, retrieve a passage that addresses, solves, or mitigates that problem.}
    \item \broadrel{}: \emph{Given a specific scientific statement, retrieve a passage that generalizes it into a broader claim, principle, or framing.}
    \item \specrel{}: \emph{Given a general scientific statement, retrieve a passage that makes it more concrete through a specific instance, detail, or special case.}
\end{itemize}
Qwen3-Embedding-0.6B and Stella-en-1.5B-v5 receive the query in their documented instruction formats, and ReasonIR-8B receives the instruction through its instruction argument; candidates are encoded without an instruction. Surprisingly, the instructions help little (Table~\ref{tab:results_silver_full}): they raise the MRR@10 of Qwen3-Embedding-0.6B by 1.0 to 2.6 points, and they change that of ReasonIR-8B by $-2.2$ to $+0.4$ points and that of Stella-en-1.5B-v5 by $-2.9$ to $+1.2$ points relative to its recommended prefix. All three models remain below every fine-tuned retriever on every move. In preliminary experiments on earlier versions of the data, we fine-tuned each move's retriever with and without a move instruction, and MRR@10 differed by at most 1.3 points.

\paragraph{Baselines}
We run three BM25 variants, which use unigrams (BM25-unigram), unigrams and bigrams (BM25-bigram), and unigrams and trigrams (BM25-trigram). All three variants remove English stopwords and apply Snowball stemming. We implement BM25 with BM25S \citep{L2024BM25SOO} and use a Faiss index \citep{faiss2017,douze2024faiss} for dense retrieval.

ReasonIR-8B runs in bf16 with a maximum length of 512 tokens. The three ColBERT models (answerai-colbert-small-v1, GTE-ModernColBERT-v1, and Reason-ModernColBERT) run in PyLate \citep{chaffin2025pylate} with MaxSim scoring, which matches each query token to its most similar candidate token and sums these similarities over the query tokens. The maximum query length is 128 tokens, and the maximum candidate length is 300 tokens. We kept PyLate's default query expansion; 99\% of the queries have at most 79 words.

\paragraph{Optimization}
For each of the 15 combinations of model, prefix, and move, we manually tried three to five hyperparameter configurations, starting from published recommendations (Table~\ref{tab:hpo}). We evaluated each run after every epoch by validation MRR@10 and restored its best checkpoint. For each combination, we likewise selected the best run by validation MRR@10. We used the test set only once, for the final results. We fine-tuned the \broadrel{} retrievers on a single GPU and the \specrel{} and \addrrel{} retrievers on four NVIDIA L40S GPUs with distributed data-parallel training. In the four-GPU runs, the in-batch negatives of each query were gathered from all four GPUs. Training took 1{,}800 GPU hours in total, and evaluation and retrieval took another 282 hours. We tracked runs with Weights~\&~Biases \citep{wandb}.

\paragraph{LLM judge}
We use GPT-5.4 as the LLM judge of the silver filter. We query it through the OpenAI Batch API, which halves the per-token cost relative to synchronous calls. Each request places the full judging prompt (Appendix~\ref{app:prompts}) in the system message and the query, marker, and gold sentence in the user message. Responses are constrained to JSON (\texttt{response\_format=json\_object}), with \texttt{reasoning\_effort=medium} and \texttt{verbosity=low}.

\paragraph{LLM reranking}
GPT-5-mini reranks the top 100 candidates of all-mpnet-base-v2 (generic prefix), ModernBERT-embed-large, and Stella-en-1.5B-v5 (recommended prefixes), before and after fine-tuning, through the Batch API with \texttt{reasoning\_effort=medium}. The system message contains the definition of the move and the rules (Appendix~\ref{app:reranker-prompt}). The user message contains the query and the 100 candidates, numbered in their first-stage order, and the prompt states that this order carries no information. The model returns JSON with the indices of its 10 best candidates. Candidates below rank 10 keep their first-stage order, so M@100 is not reported for reranked rows. Queries whose gold sentence is outside the first-stage top 100 are not sent to the reranker, because their top-10 metrics cannot change.

\begin{table}[!tp]
\centering
\scriptsize
\setlength{\tabcolsep}{3pt}
\begin{tabular}{llccll}
\toprule
Model & \# & Ep & Batch & LR & Other / Rationale \\
\midrule
\multicolumn{6}{l}{\textbf{\broadrel}}\\
\multirow{5}{*}{mpnet}
 & 0 & 4 & 64 & 2e-5 & -- \,/\, best-so-far \\
 & 1 & 3 & 64 & 2e-5 & wd .01 \,/\, published \\
 & 2 & 5 & 64 & 3e-5 & -- \,/\, higher capacity \\
 & 3 & 4 & 16 & 1e-5 & wd .01, frz .15 \,/\, anti-overfit \\
 & 4 & 4 & 32 & 1e-5 & -- \,/\, best-practice \\
\cmidrule(l){1-6}
\multirow{5}{*}{ModernBERT}
 & 0 & 3 & 32 & 2e-5 & -- \,/\, best-so-far \\
 & 1 & 1 & 32 & 2e-5 & wd .01, wu .06 \,/\, Nomic recipe \\
 & 2 & 5 & 32 & 3e-5 & -- \,/\, higher capacity \\
 & 3 & 4 & 16 & 1e-5 & wd .01, frz .15 \,/\, anti-overfit \\
 & 4 & 3 & 32 & 8e-5 & mgn .5 \,/\, probe \\
\cmidrule(l){1-6}
\multirow{5}{*}{Stella}
 & 0 & 5 & 16 & 2e-5 & mgn .5 \,/\, higher capacity \\
 & 1 & 5 & 16 & 5e-6 & frz .75 \,/\, anti-overfit \\
 & 2 & 2 & 16 & 1e-5 & mgn .5 \,/\, best-so-far \\
 & 3 & 2 & 16 & 2e-6 & -- \,/\, reference \\
 & 4 & 1 & 4  & 5e-5 & wu .05 \,/\, reference \\
\midrule
\multicolumn{6}{l}{\textbf{\addrrel}}\\
\multirow{4}{*}{mpnet}
 & 0 & 1 & 32 & 3e-5 & -- \,/\, prior winner \\
 & 1 & 2 & 32 & 3e-5 & -- \,/\, prior winner, more epochs \\
 & 2 & 1 & 16 & 2e-5 & wd .01 \,/\, published \\
 & 3 & 1 & 32 & 1e-5 & -- \,/\, probe \\
\cmidrule(l){1-6}
\multirow{4}{*}{ModernBERT}
 & 0 & 1 & 24 & 3e-5 & -- \,/\, prior winner \\
 & 1 & 2 & 24 & 3e-5 & -- \,/\, prior winner, more epochs \\
 & 2 & 1 & 16 & 8e-5 & mgn .5 \,/\, probe \\
 & 3 & 1 & 16 & 2e-5 & wd .01, wu .06 \,/\, Nomic recipe \\
\cmidrule(l){1-6}
\multirow{4}{*}{Stella}
 & 0 & 1 & 16 & 2e-6 & -- \,/\, prior winner \\
 & 1 & 2 & 16 & 2e-6 & -- \,/\, prior winner, more epochs \\
 & 2 & 1 & 16 & 2e-5 & mgn .5 \,/\, improve \\
 & 3 & 1 & 4  & 5e-5 & wu .05 \,/\, reference \\
 \midrule
\multicolumn{6}{l}{\textbf{\specrel}}\\
\multirow{3}{*}{mpnet}
 & 0 & 1 & 64 & 2e-5 & -- \,/\, prior regime \\
 & 1 & 1 & 16 & 2e-5 & wd .01 \,/\, published \\
 & 2 & 1 & 32 & 3e-5 & -- \,/\, improve \\
\cmidrule(l){1-6}
\multirow{3}{*}{ModernBERT}
 & 0 & 1 & 24 & 2e-5 & -- \,/\, prior regime \\
 & 1 & 1 & 24 & 3e-5 & -- \,/\, improve \\
 & 2 & 1 & 16 & 2e-5 & wd .01, wu .06 \,/\, Nomic recipe \\
\cmidrule(l){1-6}
\multirow{3}{*}{Stella}
 & 0 & 1 & 16 & 5e-6 & wd .01, wu .05, mgn .5 \,/\, GTE recipe \\
 & 1 & 1 & 8  & 2e-5 & wd .01, wu .05, mgn .5 \,/\, improve \\
 & 2 & 1 & 4  & 5e-5 & wu .05 \,/\, reference \\
\bottomrule
\end{tabular}
\caption{Hyperparameter configurations tried for each move and model. \#: configuration number; Ep: epochs; Batch: batch size per GPU; LR: learning rate; Other: further settings, where wd = weight decay, wu = warmup ratio, mgn = maximum gradient norm, and frz = freeze ratio; Rationale: why we tried the configuration. The number of in-batch candidates per query is the batch size times the number of GPUs (one for \broadrel{}, four for \addrrel{} and \specrel{}).}
\label{tab:hpo}
\end{table}
 
\section{Full Retrieval Results}
\label{app:full-results}

Table~\ref{tab:results_main_test_set} reports results on the unfiltered test set for BM25 and the three fine-tuned models before and after fine-tuning, and Table~\ref{tab:results_silver_full} reports results on the silver test set for every move and setup. Compared with Table~\ref{tab:results_silver} in the main text, Table~\ref{tab:results_silver_full} adds the generic-prefix setups, BM25 with bigrams and trigrams, the reranked Baselines, and Qwen3-Embedding-0.6B, ReasonIR-8B, and Stella-en-1.5B-v5 with move-specific instructions.

\paragraph{Metrics}
Each query has a single gold sentence, so Precision@$k$ equals Recall@$k$ divided by $k$, and nDCG@$k$ depends only on the rank of the gold sentence. We therefore report Recall@$k$ and MRR@$k$, with MRR@10 as the primary metric. We also report MRR@100, which exceeds MRR@10 by at most $0.8$ points for every first-stage system in Tables~\ref{tab:results_main_test_set} and~\ref{tab:results_silver_full}.

For a set of queries $Q$, where $r_q$ is the rank of the gold sentence of query $q$,
\begin{equation}
\mathrm{Recall@}k
= \frac{1}{|Q|} \sum_{q \in Q} \mathbbm{1}[r_q \leq k],
\end{equation}
which equals Accuracy@$k$, and
\begin{equation}
\mathrm{MRR@}k
= \frac{1}{|Q|} \sum_{q \in Q} \frac{\mathbbm{1}[r_q \leq k]}{r_q}.
\end{equation}

\paragraph{Unfiltered test set} The unfiltered test set contains the mined test pairs without LLM filtering. On this set, the fine-tuned ModernBERT-embed-large (recommended prefix) is the best retriever on every move: it reaches MRR@10 of 39.1 on \specrel{}, 21.5 on \addrrel{}, and 22.2 on \broadrel{} (Table~\ref{tab:results_main_test_set}), and it ranks the gold sentence in its top 10 for 56.7\%, 35.4\%, and 34.6\% of the queries. Fine-tuning improves every model on this set as well. As on the silver set, absolute scores remain low.

\paragraph{Input prefix} On the silver set, the fine-tuned ModernBERT-embed-large with the generic prefix differs from the one with the recommended prefix by at most 2.7 points on any move and metric (Table~\ref{tab:results_silver_full}).

\paragraph{LLM reranking} On \broadrel{}, the gain from reranking depends on the retriever: under one MRR@10 point for ModernBERT-embed-large, but 6.0 and 8.6 points for all-mpnet-base-v2 and Stella-en-1.5B-v5. Reranking costs an LLM call over 100 candidates for every query and cannot recover a gold sentence that the retriever missed. Fine-tuning remains useful with reranking: for every retriever, Tuned + rerank exceeds Baseline + rerank by 8.3 to 18.0 MRR@10 points on \specrel{} and \addrrel{} and by 2.4 to 3.9 points on \broadrel{} (Table~\ref{tab:results_silver_full}).

\begin{table*}[!t]
  \centering
  \small
  \setlength{\tabcolsep}{3pt}
  \renewcommand{\arraystretch}{1.2}
  \resizebox{\textwidth}{!}{
  \begin{tabular}{ll|cccc|cccc|cccc}
    \hline
    & & \multicolumn{4}{c}{\textbf{\specrel{}}} & \multicolumn{4}{c}{\textbf{\addrrel{}}} & \multicolumn{4}{c}{\textbf{\broadrel{}}} \\
    \cline{3-6}\cline{7-10}\cline{11-14}
    \textbf{System} & \textbf{Setup} & R@1 & R@10 & M@10 & M@100 & R@1 & R@10 & M@10 & M@100 & R@1 & R@10 & M@10 & M@100 \\
    \hline
    \multicolumn{14}{l}{\textit{Lexical baselines}} \\
    BM25-unigram  & --- & 10.6 & 23.3 & 14.6 & 15.1 & 5.0  & 14.0 & 7.7  & 8.1  & 6.5  & 14.2 & 8.8  & 9.2 \\
    BM25-bigram   & --- & 8.4  & 19.8 & 11.9 & 12.4 & 4.4  & 12.7 & 6.9  & 7.3  & 5.7  & 13.6 & 8.0  & 8.3 \\
    BM25-trigram  & --- & 7.0  & 18.7 & 10.5 & 11.0 & 3.7  & 11.9 & 6.1  & 6.5  & 5.0  & 13.1 & 7.3  & 7.6 \\
    \hline
    \multirow{2}{*}{\shortstack{all-mpnet-\\base-v2}}
      & generic-Baseline & 10.0 & 22.7 & 13.9 & 14.4 & 5.7  & 15.5 & 8.6  & 9.1  & 9.8  & 20.4 & 13.1 & 13.6 \\
      & generic-Tuned    & 25.2 & 48.9 & 32.8 & 33.6 & 11.3 & 28.3 & 16.6 & 17.3 & 13.4 & 26.8 & 17.7 & 18.2 \\
    \hline
    \multirow{4}{*}{\shortstack{ModernBERT-\\embed-large}}
      & generic-Baseline     & 15.1 & 32.1 & 20.4 & 21.0 & 4.0  & 11.9 & 6.4  & 6.8  & 11.0 & 23.5 & 14.9 & 15.4 \\
      & generic-Tuned        & \textbf{30.5} & 56.5 & 39.0 & \textbf{39.8} & 14.8 & \textbf{35.5} & 21.4 & \textbf{22.2} & 15.8 & 34.4 & 21.7 & 22.3 \\
    \cline{2-14}
      & recommended-Baseline & 16.3 & 33.0 & 21.7 & 22.2 & 6.5  & 16.8 & 9.7  & 10.1 & 11.8 & 23.3 & 15.5 & 16.0 \\
      & recommended-Tuned    & \textbf{30.5} & \textbf{56.7} & \textbf{39.1} & \textbf{39.8} & \textbf{15.0} & 35.4 & \textbf{21.5} & \textbf{22.2} & \textbf{16.4} & \textbf{34.6} & \textbf{22.2} & \textbf{22.7} \\
    \hline
    \multirow{4}{*}{\shortstack{Stella-\\en-1.5B-v5}}
      & generic-Baseline     & 13.6 & 29.7 & 18.6 & 19.2 & 6.9  & 19.4 & 10.8 & 11.3 & 11.8 & 26.7 & 16.4 & 16.9 \\
      & generic-Tuned        & 26.8 & 51.9 & 35.0 & 35.7 & 11.4 & 29.8 & 17.2 & 17.9 & 12.6 & 27.6 & 17.4 & 18.0 \\
    \cline{2-14}
      & recommended-Baseline & 16.3 & 34.2 & 21.9 & 22.5 & 6.8  & 18.9 & 10.5 & 11.0 & 11.8 & 25.9 & 16.1 & 16.6 \\
      & recommended-Tuned    & 27.8 & 53.7 & 36.3 & 37.0 & 12.2 & 31.1 & 18.1 & 18.8 & 12.7 & 27.2 & 17.1 & 17.7 \\
      \hline
  \end{tabular}
  }
  \caption{\label{tab:results_main_test_set}
    Results on the unfiltered \textbf{test} set (\%), per move. R@$k$ = Recall@$k$, M@$k$ = MRR@$k$; M@10 is the primary metric. Each fine-tuned model appears before (Baseline) and after (Tuned) fine-tuning, with the generic prefix and, for ModernBERT-embed-large and Stella-en-1.5B-v5, also with the recommended prefix. All systems rank the candidate pool shared by the three moves. Bold: best result in each column.
  }
\end{table*}

\begin{table*}[!t]
  \centering
  \small
  \setlength{\tabcolsep}{3pt}
  \renewcommand{\arraystretch}{1.2}
  \resizebox{\textwidth}{!}{
  \begin{tabular}{ll|cccc|cccc|cccc}
    \hline
    & & \multicolumn{4}{c}{\textbf{\specrel{}}} & \multicolumn{4}{c}{\textbf{\addrrel{}}} & \multicolumn{4}{c}{\textbf{\broadrel{}}} \\
    \cline{3-6}\cline{7-10}\cline{11-14}
    \textbf{System} & \textbf{Setup} & R@1 & R@10 & M@10 & M@100 & R@1 & R@10 & M@10 & M@100 & R@1 & R@10 & M@10 & M@100 \\
    \hline
    \multicolumn{14}{l}{\textit{\textbf{Lexical baselines}}} \\
    BM25-unigram  & --- & 13.0 & 27.3 & 17.6 & 18.1 & 5.3  & 15.2 & 8.3  & 8.7  & 7.9  & 17.6 & 10.9 & 11.3 \\
    BM25-bigram   & --- & 10.6 & 23.5 & 14.6 & 15.1 & 4.7  & 13.4 & 7.4  & 7.8  & 7.9  & 16.4 & 10.4 & 10.7 \\
    BM25-trigram  & --- & 9.0  & 22.5 & 13.0 & 13.5 & 4.0  & 12.5 & 6.5  & 6.9  & 6.5  & 15.9 & 9.3  & 9.6 \\
    \hline
    \multicolumn{14}{l}{\textit{\textbf{Dense encoders, baseline, fine-tuned, and reranked}}} \\
    \multirow{4}{*}{\shortstack{all-mpnet-base-v2}}
      & generic-Baseline            & 12.3 & 26.6 & 16.9 & 17.5 & 6.1  & 16.5 & 9.2  & 9.7  & 11.6 & 24.3 & 15.6 & 16.1 \\
      & generic-Tuned               & 28.5 & 54.3 & 37.0 & 37.8 & 12.8 & 31.7 & 18.7 & 19.4 & 16.3 & 32.5 & 21.8 & 22.4 \\
      & generic-Baseline + rerank   & 30.8 & 40.4 & 34.4 & --   & 19.8 & 27.7 & 22.7 & --   & 21.4 & 33.2 & 24.9 & --   \\
      & generic-Tuned + rerank      & 45.3 & 66.0 & 52.4 & --   & 28.5 & 45.9 & 34.2 & --   & 23.8 & 37.7 & 27.8 & --   \\
    \hline
    \multirow{6}{*}{\shortstack{ModernBERT-embed-large}}
      & generic-Baseline     & 18.5 & 38.2 & 24.9 & 25.5 & 3.9  & 11.8 & 6.3  & 6.7  & 13.2 & 27.4 & 17.6 & 18.2 \\
      & generic-Tuned        & 34.9 & 62.8 & 44.1 & 44.8 & 16.9 & \textbf{40.2} & 24.4 & 25.2 & 19.3 & 40.6 & 26.2 & 26.9 \\
    \cline{2-14}
      & recommended-Baseline          & 20.5 & 39.1 & 26.7 & 27.3 & 6.7  & 17.8 & 10.2 & 10.6 & 13.4 & 26.0 & 17.3 & 17.9 \\
      & recommended-Tuned             & \textbf{37.2} & \textbf{65.5} & \textbf{46.7} & \textbf{47.4} & \textbf{17.2} & 40.0 & \textbf{24.5} & \textbf{25.3} & \textbf{20.9} & \textbf{41.4} & \textbf{27.3} & \textbf{27.8} \\
      & recommended-Baseline + rerank & 37.0 & 49.9 & 41.7 & --   & 20.9 & 28.9 & 23.8 & --   & 21.6 & 34.6 & 25.8 & --   \\
      & recommended-Tuned + rerank    & \underline{47.5} & \underline{69.9} & \underline{55.1} & --   & \underline{30.8} & \underline{52.1} & \underline{37.7} & --   & 22.3 & \underline{40.9} & 28.2 & --   \\
    \hline
    \multirow{7}{*}{\shortstack{Stella-en-1.5B-v5}}
    & generic-Baseline     & 16.5 & 34.9 & 22.5 & 23.0 & 7.3  & 20.5 & 11.4 & 11.9 & 14.7 & 31.5 & 19.7 & 20.3 \\
      & generic-Tuned        & 31.0 & 57.7 & 39.8 & 40.5 & 12.9 & 33.0 & 19.2 & 20.0 & 15.8 & 33.4 & 21.7 & 22.4 \\
    \cline{2-14}
      & recommended-Baseline          & 20.0 & 39.8 & 26.4 & 27.0 & 7.2  & 20.1 & 11.0 & 11.6 & 13.7 & 29.6 & 18.6 & 19.2 \\
      & recommended-Tuned             & 31.8 & 60.0 & 41.0 & 41.8 & 13.8 & 34.4 & 20.3 & 21.0 & 15.8 & 31.8 & 20.9 & 21.7 \\
      & recommended-Baseline + rerank & 38.2 & 52.1 & 43.1 & --   & 22.4 & 32.8 & 26.2 & --   & 21.1 & 35.3 & 25.6 & --   \\
      & recommended-Tuned + rerank    & 45.6 & 68.0 & 53.4 & --   & 28.5 & 47.1 & 34.5 & --   & \underline{24.0} & \underline{40.9} & \underline{29.5} & --   \\
    \cline{2-14}
      & per-move-Baseline             & 17.4 & 36.2 & 23.5 & 24.1 & 8.0  & 21.4 & 12.2 & 12.8 & 14.2 & 30.5 & 19.2 & 20.0 \\
    \hline
    \multicolumn{14}{l}{\textit{\textbf{Zero-shot baselines}}} \\
    EmbeddingGemma-300m       & Baseline & 22.1 & 42.3 & 28.8 & 29.4 & 4.9 & 14.6 & 7.8  & 8.3  & 14.0 & 28.9 & 18.5 & 19.1 \\
    \multirow{2}{*}{Qwen3-Embedding-0.6B}
      & generic-Baseline  & 17.6 & 37.0 & 24.0 & 24.5 & 6.8 & 18.9 & 10.4 & 10.9 & 12.0 & 27.1 & 16.5 & 17.1 \\
      & per-move-Baseline & 19.0 & 38.4 & 25.3 & 25.9 & 8.6 & 22.9 & 13.0 & 13.6 & 13.0 & 26.7 & 17.5 & 18.2 \\
    \multirow{2}{*}{ReasonIR-8B}
      & generic-Baseline  & 22.6 & 44.3 & 29.6 & 30.2 & 8.0 & 21.7 & 12.1 & 12.7 & 13.0 & 30.8 & 18.3 & 18.9 \\
      & per-move-Baseline & 20.9 & 42.5 & 27.7 & 28.4 & 6.3 & 18.4 & 9.9  & 10.5 & 13.9 & 29.5 & 18.7 & 19.4 \\
    \cline{2-14}
    answerai-colbert-small-v1 & MaxSim-Baseline & 15.6 & 30.7 & 20.5 & 21.0 & 4.2 & 11.0 & 6.3  & 6.6  & 7.4  & 15.8 & 10.0 & 10.3 \\
    GTE-ModernColBERT-v1      & MaxSim-Baseline & 17.4 & 34.3 & 22.7 & 23.2 & 5.5 & 14.9 & 8.4  & 8.8  & 7.5  & 20.9 & 11.9 & 12.4 \\
    Reason-ModernColBERT      & MaxSim-Baseline & 17.2 & 34.0 & 22.6 & 23.1 & 5.9 & 16.7 & 9.2  & 9.7  & 9.1  & 22.1 & 13.6 & 14.1 \\
    \hline
  \end{tabular}
  }
  \caption{\label{tab:results_silver_full}
    Full results on the \textbf{silver} test set (\%), per move. R@$k$ = Recall@$k$, M@$k$ = MRR@$k$; M@10 is the primary metric. The silver set has 7{,}327 \specrel{}, 9{,}668 \addrrel{}, and 584 \broadrel{} queries. Each fine-tuned model appears before (Baseline) and after (Tuned) fine-tuning, with the generic prefix and, for ModernBERT-embed-large and Stella-en-1.5B-v5, also with the recommended prefix. All systems rank the candidate pool shared by the three moves. The three ColBERT models score candidates with MaxSim. \emph{per-move}: the query carries a move-specific instruction (Appendix~\ref{app:training-details}). \emph{+ rerank}: GPT-5-mini reranks the top 100 candidates of that retriever listwise, at medium reasoning effort, and returns the 10 best; candidates below rank 10 keep their first-stage order, and queries whose gold sentence is outside the top 100 keep their first-stage ranking. Only the all-mpnet-base-v2 generic and the ModernBERT-embed-large and Stella-en-1.5B-v5 recommended setups were reranked. Bold: best result without reranking in each column. Underline: best reranked result in each column.
  }
\end{table*}

\section{Silver Test Set and LLM Judge: Details}
\label{app:judge-details}

\paragraph{Annotators} Five graduate students in computer science labeled sentence pairs: three female and two male, all over 25, and all with research experience and experience in text annotation. We recruited them through university online channels and paid them \$30 per hour. They received the same prompts as the LLM judge. One of them, whom we call the calibration annotator, labeled 100 pairs per move for prompt selection, and all five labeled 60 further pairs per move that were held out from prompt selection. All annotators agreed to the release of their anonymized annotations; we release only the labels, without any information about the annotators.

\paragraph{Sampling} We sampled 10K \specrel{} test pairs by stratified sampling with equal allocation across markers: each marker contributes the same number of pairs, or all of its pairs if it has fewer, so that frequent markers do not dominate the sample. We took all 14,020 \addrrel{} and all 1,410 \broadrel{} test pairs.

\paragraph{Judging criteria} Each judging prompt checks the query and the gold sentence separately for grammar, coherence, and self-containment. A sentence is self-contained if it can be understood without the rest of the paper; the gold sentence may refer back to the query. Each prompt also checks the relation between the two sentences. For \addrrel{}, the query must state a problem, and the gold sentence must directly address that problem. For \broadrel{} and \specrel{}, the two sentences must share their main topic, not just part of it. A \broadrel{} gold sentence may be at a higher level of abstraction than its query, but a pair is dropped if the main themes of the two sentences are not clearly aligned. For both \broadrel{} and \specrel{}, the prompts also check the direction of the relation. A \specrel{} gold sentence must be a specific case of the query, not merely something lower in a hierarchy. Pairs are dropped when the roles are reversed, or when the gold sentence continues the topic at the same level of abstraction without specifying or broadening it. Appendix~\ref{sec:silver_prompts_classifier} lists the prompts.

\paragraph{Prompt selection and agreement} The labels of a single prompt depend on its wording, so a prompt can be too strict on some pairs and too lenient on others. For each move, we therefore wrote four to six prompt variants, compared their labels with the calibration annotator's labels on 100 pairs of that move (300 pairs in total), and kept the two variants with the highest $F_1$. A pair must be accepted by both prompts. On the 100 pairs per move used for selection, the retained prompts reach $F_1$ of .86--.88 for \specrel{}, .85 for \addrrel{}, and .77--.79 for \broadrel{} against the calibration annotator (Table~\ref{tab:prompt-agreement}, right). On the 60 held-out pairs per move labeled by all five annotators, the average $F_1$ between two annotators is .90 for \specrel{}, .87 for \addrrel{}, and .82 for \broadrel{} (Table~\ref{tab:prompt-agreement}, left). On the same pairs, the retained prompts reach an average $F_1$ against a single annotator of .87--.91 for \specrel{}, .73--.76 for \addrrel{}, and .75--.81 for \broadrel{}. Against the majority label of the panel, the retained prompts reach $F_1$ of .88, .83, and .76 for \specrel{}, \addrrel{}, and \broadrel{}, and .96, .82, and .86 on the pairs on which the panel is unanimous. On the panel, the combined rule (\emph{Both (AND)}) agrees with the annotators about as well as the individual retained prompts, and for \specrel{} and \broadrel{} about as well as two annotators agree with each other (Table~\ref{tab:prompt-agreement}, left).

\paragraph{Filtering} We applied GPT-5.4 with the two retained prompts of each move to the sampled test pairs. The silver filter sees the query, the marker, and the gold sentence, in the form \texttt{querySentence. discourse-marker, goldSentence}, and we keep a pair only if both prompts accept it. The resulting silver set has 17,579 queries: 7,327 for \specrel{}, 9,668 for \addrrel{}, and 584 for \broadrel{}. A pair rejected by the filter is removed from the set of queries, but its gold sentence stays in the candidate pool. Silver queries are ranked against the full test pool of 404,371 sentences, which includes the distractors and contains no training sentence.

\paragraph{Anaphoric \addrrel{} markers} The markers of 93.4\% of the \addrrel{} test pairs contain an explicit anaphor (\emph{this}, \emph{these}, \emph{such}, \emph{it}, or \emph{the above}), as in \emph{To address this issue}. Such markers occur in only three of the sampled \specrel{} pairs and two of the \broadrel{} pairs. Within \addrrel{}, the silver filter accepts 71.9\% of the pairs whose marker contains an explicit anaphor but only 27.7\% of the pairs whose marker does not, an odds ratio of 6.7 (Fisher's exact test, $p<.001$). This effect is specific to \addrrel{}.

\begin{table*}[!t]
\centering
\small
\begin{tabular}{ll rr | rrrrr}
\toprule
& & \multicolumn{2}{c}{Annotator panel} & \multicolumn{5}{|c}{Single annotator} \\
\cmidrule(lr){3-4}\cmidrule(lr){5-9}
Move & Source & Agr. & $F_1$ & Agr. & $F_1$ & $\kappa$ & P & R \\
\midrule
\multirow{4}{*}{\addrrel{}}
 & Human annotators & 83.3 & .87 & -- & -- & -- & -- & -- \\
 & Minimal-v2       & 75.0 & .76 & 82.0 & .85 & .62 & .83 & .88 \\
 & Ordered-v2       & 71.7 & .73 & 81.0 & .85 & .60 & .82 & .88 \\
 & Both (AND)       & 76.7 & .76 & 81.0 & .84 & .60 & .84 & .85 \\
\midrule
\multirow{4}{*}{\textsc{Specify}}
 & Human annotators & 86.7 & .90 & -- & -- & -- & -- & -- \\
 & Anchor-full-v2   & 86.7 & .91 & 82.0 & .88 & .52 & .84 & .93 \\
 & Anchor-compact   & 80.0 & .87 & 80.0 & .86 & .48 & .83 & .90 \\
 & Both (AND)       & 85.0 & .89 & 82.0 & .88 & .55 & .86 & .89 \\
\midrule
\multirow{4}{*}{\textsc{Broaden}}
 & Human annotators & 83.3 & .82 & -- & -- & -- & -- & -- \\
 & Directional      & 80.0 & .81 & 74.0 & .77 & .47 & .85 & .71 \\
 & Relaxed          & 70.0 & .75 & 73.0 & .79 & .42 & .77 & .81 \\
 & Both (AND)       & 80.0 & .81 & 73.0 & .76 & .46 & .86 & .68 \\
\bottomrule
\end{tabular}
\caption{Agreement of the retained prompts with human labels. Left: 60 pairs per move labeled by a panel of five annotators. Scores are averaged over pairs of annotators (human rows) or over prompt--annotator pairs (prompt rows), so the rows are directly comparable. Right: 300 pairs (100 per move) labeled by the single annotator on whom the prompts were calibrated. We report P, R, and $\kappa$ only against the single annotator (right). In the right panel, the human rows are empty (dashes), because agreement between humans requires at least two annotators. \emph{Both (AND)}: the conjunction of the two retained prompts---the rule the released silver set applies, keeping a pair only if both prompts accept it; its panel columns are scored against the annotators exactly as the prompt rows are. Agr.: percentage of pairs with the same label. The prompts are Listings~\ref{lst:silver_prompts_classifier_address_p1} (Minimal-v2), \ref{lst:silver_prompts_classifier_address_p2} (Ordered-v2), \ref{lst:silver_prompts_classifier_specify_p1} (Anchor-full-v2), \ref{lst:silver_prompts_classifier_specify_p2} (Anchor-compact), \ref{lst:silver_prompts_classifier_broaden_p1} (Relaxed), and \ref{lst:silver_prompts_classifier_broaden_p2} (Directional).}
\label{tab:prompt-agreement}
\end{table*}

\section{Extended Analyses}
\label{app:analysis-details}

\subsection{Lexical Overlap}
For each query, we computed the mean word-level Jaccard similarity between the query and its top-$k$ candidates and compared it with the Jaccard similarity between the query and its gold sentence. Before computing overlap, we lowercased the text, tokenized it on word characters, removed English stopwords, and applied Snowball stemming, as in our BM25 setup.

\begin{figure}
    \centering
    \includegraphics[width=1\linewidth]{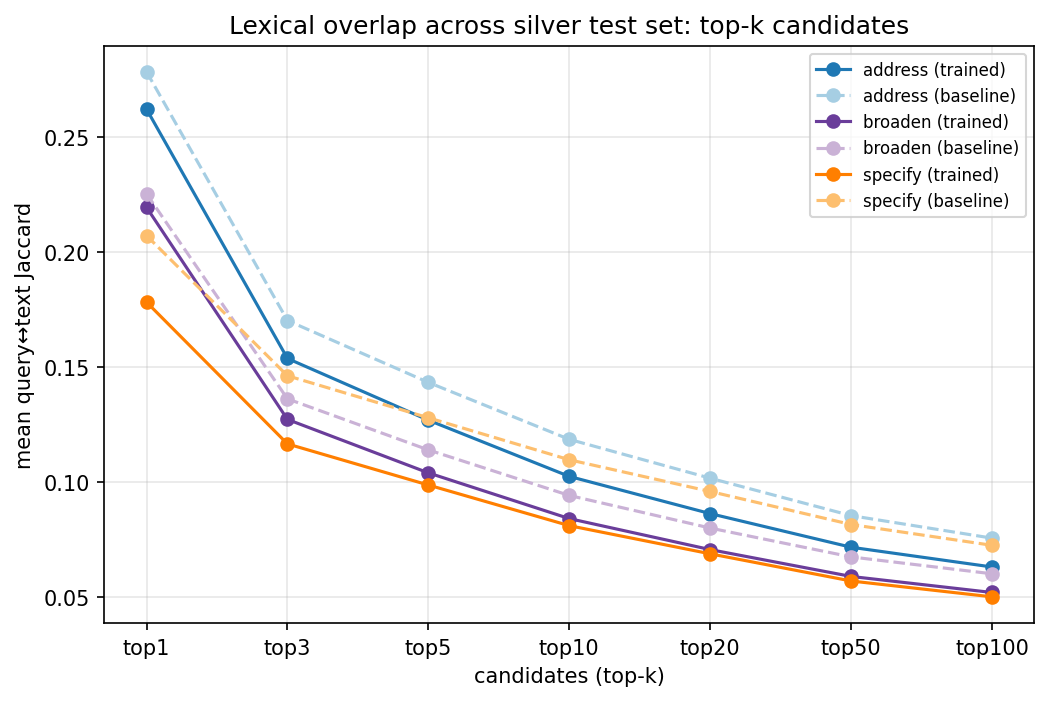}
    \caption{Mean lexical overlap between a silver query and its top-$k$ candidates, for ModernBERT-embed-large before (baseline) and after (trained) fine-tuning on each move. Overlap is the Jaccard similarity of the two sentences' unigrams after stopword removal and stemming.}
    \label{fig:query-candidate_jaccard}
\end{figure}

The candidates of the fine-tuned ModernBERT-embed-large overlap less with the query than those of its baseline in all 24 comparisons, which cover four values of $k$ ($k \in \{1,3,5,10\}$), three moves, and two query sets (the unfiltered test set and the silver set). On \specrel{}, the fine-tuned retriever's top-10 candidates overlap with the query even less than the gold sentence does, by $.010$ on the unfiltered test set and $.019$ on the silver set. Figure~\ref{fig:query-candidate_jaccard} shows the overlap on the silver set for seven cutoffs, from $k=1$ to $k=100$.

\subsection{Oracle Move Filtering}
\label{app:oracle-filtering}
A fine-tuned retriever could succeed by a shortcut: it could recognize sentences written in the style of the requested move and then pick the one closest in topic to the query. To test this explanation, we restrict the candidate pool to gold sentences of the queried move. We then evaluate ModernBERT-embed-large (recommended prefix), before and after fine-tuning, on the same silver queries as in the main experiments. We use two restricted pools (Table~\ref{tab:results_silver_oracle_pools}): (a) the gold sentences of all test pairs of the move, and (b) only the gold sentences of its silver pairs.

\noindent\textbf{A style shortcut does not explain the gains.}
In each restricted pool, every candidate has the style of the queried move, yet fine-tuning still adds 22.1, 19.4, and 14.8 MRR@10 points for \specrel{}, \addrrel{}, and \broadrel{} in pool (a), and 26.6, 20.1, and 15.8 points in pool (b). A style shortcut alone cannot produce gains of this size. Scores are not comparable across pools of different sizes, so each panel of Table~\ref{tab:results_silver_oracle_pools} should be read on its own. Because the baseline scores much higher in the small pools than in the full test pool shared by all moves, the relative gains are smaller there (for \addrrel{}, $1.7\times$ in pool (a) instead of $2.4\times$ in the full pool).

\begin{table*}[h]
  \centering
  \small
  \setlength{\tabcolsep}{3pt}
  \renewcommand{\arraystretch}{1.2}
  \resizebox{\textwidth}{!}{
  \begin{tabular}{ll|cccc|cccc|cccc}
    \hline
    & & \multicolumn{4}{c}{\textbf{\specrel{}}} & \multicolumn{4}{c}{\textbf{\addrrel{}}} & \multicolumn{4}{c}{\textbf{\broadrel{}}} \\
    \cline{3-6}\cline{7-10}\cline{11-14}
    \textbf{Candidate pool} & \textbf{Setup} & R@1 & R@10 & M@10 & M@100 & R@1 & R@10 & M@10 & M@100 & R@1 & R@10 & M@10 & M@100 \\
    \hline
    \multirow{2}{*}{\shortstack[l]{(a) Move's test\\ gold sentences}}
      & Baseline & 29.5 & 48.0 & 35.4 & 36.0 & 23.1 & 41.7 & 28.7 & 29.4 & 39.7 & 63.5 & 46.9 & 47.8 \\
      & Tuned    & \textbf{48.8} & \textbf{74.7} & \textbf{57.5} & \textbf{58.1} & \textbf{40.3} & \textbf{65.4} & \textbf{48.1} & \textbf{48.9} & \textbf{54.8} & \textbf{77.6} & \textbf{61.7} & \textbf{62.4} \\
    \hline
    \multirow{2}{*}{\shortstack[l]{(b) Move's silver\\ gold sentences}}
      & Baseline & 46.0 & 64.9 & 52.0 & 52.6 & 25.4 & 44.7 & 31.1 & 31.8 & 44.7 & 69.7 & 52.6 & 53.5 \\
      & Tuned    & \textbf{72.0} & \textbf{90.8} & \textbf{78.6} & \textbf{79.0} & \textbf{43.4} & \textbf{68.4} & \textbf{51.2} & \textbf{52.0} & \textbf{61.3} & \textbf{84.6} & \textbf{68.4} & \textbf{69.0} \\
    \hline
  \end{tabular}
  }
  \caption{\label{tab:results_silver_oracle_pools}
    \textbf{Oracle move filtering.} Results on the \textbf{silver} test set (\%), per move, when the candidate pool contains only candidates of the queried move. All rows are ModernBERT-embed-large with the recommended (\texttt{search\_query}) prefix, before (\emph{Baseline}) and after (\emph{Tuned}) fine-tuning on the move. R@$k$ = Recall@$k$, M@$k$ = MRR@$k$; M@10 is the primary metric. For \specrel{}, \addrrel{}, and \broadrel{}, both panels use the same 7{,}327, 9{,}668, and 584 silver queries and differ only in the candidate pool. Pool (a) contains the gold sentences of all test pairs of the move (94{,}079, 14{,}020, and 1{,}410 candidates). Pool (b) contains only the gold sentences of the move's silver pairs (7{,}327, 9{,}668, and 584 candidates). Bold: the better of Baseline and Tuned within each panel.
  }
\end{table*}

\subsection{Cross-Move Transfer}
\label{app:cross-move}
A fine-tuned retriever could also improve by learning a skill common to all three moves, such as finding the sentence that follows the query in its paper, rather than the move itself. Such a skill would transfer from one move to another. To test this, we evaluate the retrievers fine-tuned on \specrel{} and on \addrrel{}, the two moves with the most training pairs (Table~\ref{tab:temporal-split-full}), on the silver queries of both moves: all-mpnet-base-v2 with the generic prefix, and ModernBERT-embed-large and Stella-en-1.5B-v5 with their recommended prefixes.

For every model, the retriever fine-tuned on the queried move scores best. Relative to the model before fine-tuning, fine-tuning on the queried move raises MRR@10 by 55\% to 119\% on \specrel{} queries and by 85\% to 140\% on \addrrel{} queries, whereas fine-tuning on the other move raises it by at most 34\% and 35\%, respectively. For example, fine-tuning ModernBERT-embed-large on \specrel{} raises its \specrel{} MRR@10 by 75\%, but fine-tuning it on \addrrel{} raises it by only 13\%. Fine-tuning thus teaches more than a skill common to all moves.

\section{Human Inspiration Evaluation}
\label{app:human-eval-top10}

\paragraph{Annotators} Five computer science graduate students, three female and two male, all over 25 and each with at least one publication, labeled the queries, 18 to 33 queries each. We recruited them through university online channels and paid them \$30 per hour. Annotation took a little over 30 hours in total. All annotators agreed to the release of their anonymized annotations; we release only the labels, without any information about the annotators.

\paragraph{Training} Each annotator received written guidelines, watched a video that explains the motivation, scope, and framing of the task, and labeled an example query together with the first author. We will release the guidelines, the video, and screenshots of the annotation interface in our GitHub repository. The guidelines frame candidates as inspirations: a candidate is useful if it would help the annotator address the problem in the query (\addrrel{}), broaden it (\broadrel{}), or find a concrete example of it (\specrel{}).

   \paragraph{Protocol} The queries come from a pool of 400 silver queries per move, sampled by marker with equal allocation; for \specrel{}, the pool was drawn from an earlier, confidence-filtered subset of 5,707 pairs, all of which remain valid under the final silver rule. Each annotator was assigned queries from their own research subfield, to simulate a researcher who seeks inspiration in their own area. For each query, the annotator saw the query, the requested move, and the top 10 candidates of ModernBERT-embed-large before and after fine-tuning as two lists labeled List A and List B, side by side. The order of the two lists was shuffled for each query, and annotators were blind to condition. The annotator marked each candidate as useful or not as an inspiration for the requested move and chose the more useful list, allowing ties; an optional comment field was available. We count a query only if all 20 candidates are labeled.
   
\paragraph{Metrics} A candidate marked useful counts as relevant. HR@10 is the fraction of queries with at least one useful candidate in the top 10, P@10 is the number of useful candidates in the top 10 divided by 10, and MRR@10 is the mean reciprocal rank of the first useful candidate. For a query labeled twice, we average the two annotators' scores. We test each difference with a paired sign-flip permutation test over queries (10{,}000 permutations).

\paragraph{Results} Table~\ref{tab:human_eval} reports the results per move. Fine-tuning improves every metric for every move ($p<.01$).

\subsection{Inspirations Beyond the Gold Sentence}
\label{app:human-beyond-gold}

\paragraph{Definitions} A candidate is \emph{beyond the gold sentence} if it is useful and is not the gold sentence. \emph{Gold-only} metrics count only the gold sentence as relevant, as on the silver set, but on the same queries and lists that the annotators saw. To find a candidate's paper, we match its text to the test corpus.

\paragraph{Gold sentence and beyond} Table~\ref{tab:human_beyond_gold} reports the results per move. Fine-tuning raises the share of queries with a useful candidate other than the gold sentence on every move. With every useful candidate counted, MRR@10 is 1.5 to 3.1 times the gold-only MRR@10 for both retrievers. These 125 queries are not a random sample of the silver set, so their gold-only scores differ from those in Table~\ref{tab:results_silver}.

\paragraph{Cross-paper candidates} Table~\ref{tab:crosspaper-app} reports where the useful candidates other than the gold sentence come from. Fine-tuning raises their cross-paper share on every move.

\begin{table}[h]
\centering\small
\setlength{\tabcolsep}{3pt}
\resizebox{\columnwidth}{!}{%
\begin{tabular}{llrrrrrr}
\toprule
Move & Setup & HR & Gold & Gold useful & Beyond & MRR & MRR gold \\
\midrule
\multirow{2}{*}{\specrel{} (39)} & Baseline & 50.0 & 25.6 & 25.6 & 39.7 & 37.4 & 17.8 \\
 & Tuned & \textbf{87.2} & \textbf{69.2} & \textbf{69.2} & \textbf{62.8} & \textbf{63.9} & \textbf{43.3} \\
\midrule
\multirow{2}{*}{\addrrel{} (40)} & Baseline & 42.5 & 12.5 & 12.5 & 37.5 & 21.7 & 7.0 \\
 & Tuned & \textbf{85.0} & \textbf{32.5} & \textbf{32.5} & \textbf{70.0} & \textbf{42.3} & \textbf{14.6} \\
\midrule
\multirow{2}{*}{\broadrel{} (46)} & Baseline & 50.0 & 23.9 & 19.6 & 47.8 & 36.8 & 17.6 \\
 & Tuned & \textbf{72.8} & \textbf{37.0} & \textbf{32.6} & \textbf{62.0} & \textbf{49.5} & \textbf{24.4} \\
\midrule
\multirow{2}{*}{All (125)} & Baseline & 47.6 & 20.8 & 19.2 & 42.0 & 32.1 & 14.3 \\
 & Tuned & \textbf{81.2} & \textbf{45.6} & \textbf{44.0} & \textbf{64.8} & \textbf{51.7} & \textbf{27.1} \\
\bottomrule
\end{tabular}%
}
\caption{Useful candidates and the gold sentence in the top 10 of ModernBERT-embed-large, for \emph{Baseline} and \emph{Tuned}, on the annotated queries (\%, mean over queries; the number of queries is in parentheses). HR and MRR: hit rate and MRR@10 with every useful candidate counted as relevant. Gold: queries whose top 10 contains the gold sentence (gold-only HR@10). Gold useful: queries whose top 10 contains the gold sentence and the annotator marked it useful. Beyond: queries whose top 10 contains a useful candidate other than the gold sentence. MRR gold: gold-only MRR@10. Bold: the better of Baseline and Tuned for each move.}
\label{tab:human_beyond_gold}
\end{table}

\begin{table}[h]
\centering\small
\setlength{\tabcolsep}{3pt}
\resizebox{\columnwidth}{!}{%
\begin{tabular}{llrrrr}
\toprule
Move & Setup & Useful & Cross \% & Papers & All \% \\
\midrule
\multirow{2}{*}{\specrel{}} & Baseline & 40 & 77.5 & 39 & 91.7 \\
 & Tuned & 56 & 87.5 & 54 & 94.1 \\
\midrule
\multirow{2}{*}{\addrrel{}} & Baseline & 23 & 87.0 & 22 & 95.8 \\
 & Tuned & 65 & 96.9 & 65 & 96.0 \\
\midrule
\multirow{2}{*}{\broadrel{}} & Baseline & 34 & 82.4 & 32 & 95.7 \\
 & Tuned & 59 & 89.8 & 57 & 95.9 \\
\midrule
\multirow{2}{*}{All} & Baseline & 97 & 81.4 & 93 & 94.5 \\
 & Tuned & 180 & 91.7 & 176 & 95.4 \\
\bottomrule
\end{tabular}%
}
\caption{Origin of useful candidates other than the gold sentence in the top 10 of ModernBERT-embed-large, on the 125 annotated queries. Useful: number of such candidates, pooled over annotations. Cross: percentage of them that come from a paper other than the query's. Papers: number of distinct papers among them. All: percentage of all non-gold top-10 candidates that come from a paper other than the query's.}
\label{tab:crosspaper-app}
\end{table}

\section{Prompts}
\label{app:prompts}

\subsection{Discourse Markers Classifier: \addrrel{}, \broadrel{}}
\label{sec:discourse_markers_classifier}
We applied this classifier only to \addrrel{} and \broadrel{}: the \specrel{} markers already yield the largest pool of pairs, so we did not expand that lexicon with LLMs, and two NLP experts vetted all of its markers.
\begin{lstlisting}[style=promptstyle,caption={\texttt{Discourse Markers Classifier: \addrrel{}}.},label={lst:discourse_markers_classifier_address}]
You are an expert linguistic judge evaluating connective phrases in scientific text. Your job is to decide, with high certainty, whether a given connective phrase reliably signals that the sentence following it offers a SOLUTION, IMPROVEMENT, or MITIGATION that is DIRECTLY inspired by and targets the problem described in the sentence preceding it.

A "problem-solution" relationship means sentence 1 describes a problem, limitation, challenge, or gap, and sentence 2 proposes an approach, method, or remedy that is directly motivated by that specific problem. This is the "to address this, we..." relationship - the solution must be a direct response to the prior problem, not merely a related contribution that happens to follow.

Return True ONLY if, across the vast majority of plausible scientific contexts, the phrase forces a reading in which the following sentence is a remedy DIRECTLY inspired by the prior problem. Return False if the phrase could plausibly introduce other relationships such as generalization, specification, consequence, paraphrase, emphasis, exemplification, contrast without a tied remedy, or continuation of the problem description.

When uncertain, return False. The bar for True is not "a solution follows" but "a solution clearly tied to the specific problem just stated follows."

Edge cases and guidance:
- The phrase must anaphorically or purposively tie the solution to the prior problem. Generic cues that do not reference the problem are False.
- Contrast phrases ("however", "on the other hand", "in contrast", "nevertheless", "yet", "nonetheless") are always False - they mark opposition without committing to a remedy tied to the prior problem.
- Causal/inferential phrases ("therefore", "thus") are False - they can mark consequence or takeaway, not reliably a targeted remedy.
- Additive phrases ("furthermore", "moreover", "additionally", "in addition") are False - they extend discourse without tying what follows to the prior problem.
- Generalization ("more generally", "more broadly"), exemplification ("for example", "for instance", "e.g."), restatement ("in other words", "that is", "i.e.", "namely"), and emphasis ("notably", "importantly", "crucially", "indeed") phrases are all False.
- Bare purposive phrases without an anaphoric tie ("to address", "to overcome", "to mitigate", "to solve", "to tackle", "to alleviate", "to circumvent", "to remedy") are False - they can head infinitival clauses for any purpose.
- Bare first-person contribution verbs without an anaphoric tie ("we propose", "we introduce", "we present", "we develop", "our approach", "our method") are False - they often open contributions cold, without direct motivation from the prior sentence.
- Anaphoric purposive phrases that explicitly reference the prior problem are True: "to address this", "to address this issue", "to address this problem", "to address this limitation", "to address these challenges", "to address the above", "to overcome this", "to overcome this limitation", "to overcome these limitations", "to mitigate this", "to tackle this issue", "to alleviate this", "to resolve this", "to solve this problem", "to circumvent this", "to remedy this".
- Explicit solution-pivots with back-reference are True: "as a solution to this", "our solution to this", "to fix this".
- Contribution verbs combined with an anaphoric tie are True (e.g., "to address this, we propose").

Output strictly valid JSON with this schema:
{
  "label": <true | false>,
  "rationale": "<one sentence explaining the decision>",
  "confidence": <"high" | "medium" | "low">
}

Do not include any text outside the JSON object. Do not wrap the JSON in markdown fences.
\end{lstlisting}

\begin{lstlisting}[style=promptstyle,caption={\texttt{Discourse Markers Classifier: \broadrel{}}.},label={lst:discourse_markers_classifier_broaden}]
You are an expert linguistic judge evaluating connective phrases in scientific text.
Your job is to decide, with high certainty, whether a given connective phrase reliably signals that the sentence following it is a GENERALIZATION of the sentence preceding it.

A "generalization" relationship means sentence 2 broadens, abstracts, or extends the scope of sentence 1 - moving from a specific claim to a wider one.
This is the "more generally" relationship.

Return True ONLY if, across the vast majority of plausible scientific contexts, the phrase forces a generalization reading.
Return False if the phrase is ambiguous and could plausibly introduce other relationships such as:
- Specification or narrowing (opposite direction)
- Implication, consequence, or takeaway
- Paraphrase or restatement
- Emphasis or highlighting
- Example or illustration
- Contrast or qualification

When uncertain, return False. We prefer to discard ambiguous connectives rather than admit noisy ones into the dataset.

Edge cases and guidance:
- Phrases like "in general" or "generally" are ambiguous because they can introduce a default/baseline claim rather than a genuine broadening. Return False for these unless the generalization reading clearly dominates.
- Phrases that combine generalization with another function (e.g., "more generally speaking, therefore") should be judged on the dominant function. If generalization is dominant and reliable, return True.
- Purely additive or continuative phrases ("furthermore", "moreover", "additionally") are False - they extend discourse but do not necessarily broaden scope.
- Causal or inferential phrases ("therefore", "thus", "hence") are False - they mark inference, not generalization.
- Exemplification phrases ("for example", "for instance", "such as", "e.g.") are False - they move from general to specific, the opposite direction.
- Restatement phrases ("in other words", "that is", "i.e.", "namely") are False - they paraphrase at the same level of abstraction.
- Contrast phrases ("however", "on the other hand", "in contrast", "nevertheless") are False - they mark opposition, not generalization.
- Emphasis phrases ("notably", "importantly", "crucially", "indeed") are False - they highlight salience, not broaden scope.
- Phrases explicitly marking scope expansion ("more generally", "more broadly", "in a broader sense", "in a more general sense", "at a higher level of abstraction") are True.

Output strictly valid JSON with this schema:
{
  "label": <true | false>,
  "rationale": "<one sentence explaining the decision>",
  "confidence": <"high" | "medium" | "low">
}

Do not include any text outside the JSON object. Do not wrap the JSON in markdown fences.
\end{lstlisting}

\subsection{Silver Test Set: Query-Gold \addrrel{}, \specrel{} and \broadrel{}}
\label{sec:silver_prompts_classifier}

\begin{lstlisting}[style=promptstyle,caption={\texttt{Silver Query sentence-Gold sentence: \addrrel{} - Prompt 1}.},label={lst:silver_prompts_classifier_address_p1}]
Input is "S1 (the problem). connecting phrase, S2 (the solution/mitigation)." Judge S1 and S2 independently, then the match.

Note: The connecting phrase is consistent with a *problem-solution* relation, but its presence does not guarantee that this is the actual relation between the sentences. Judge the relation from the sentence content itself, not from the connecting phrase alone. Ignore the connecting phrase when assessing the match.

is_problem_valid = True only if S1 states a clear, specific problem AND the core problem is articulated in S1 itself (a pronoun pointing outside S1 is OK only if it is NOT the core problem). False if garbled, if it only says a problem exists without saying what, or if the core problem hides behind an external reference. False if S1 isn't actually a problem - it's a capability, requirement, or assumption.

is_solution_valid = True only if S2 describes a specific solution/mitigation an expert would recognize from its terms, with the core solution articulated in S2 (a reference back to the problem sentence is fine). False if garbled, or if it only names/asserts a solution exists or defers details ("the following ..."). False if S2 only explains how a method works, or is a diagnostic/experimental setup, or is a consequence/derivation, not a remedy.

match = True only if both are valid AND S2 directly targets the core problem in S1.

OUTPUT (JSON only):
{"is_problem_valid":{"is_valid":"True|False","reason":"<1 sentence>"},
"is_solution_valid":{"is_valid":"True|False","reason":"<1 sentence>"},
"match_evaluation":{"is_match":"True|False","reason":"<1 sentence>"}}

\end{lstlisting}

\begin{lstlisting}[style=promptstyle,breakatwhitespace=false,caption={\texttt{Silver Query sentence-Gold sentence: \addrrel{} - Prompt 2}.},label={lst:silver_prompts_classifier_address_p2}]
Given a passage from scientific literature of the form "S1 (the problem). connecting phrase, S2 (the solution/mitigation).", evaluate whether it forms a valid problem-solution pair for an information retrieval test set. S1 is the problem, S2 is the solution/mitigation. Execute the steps in order.

Note: The connecting phrase is consistent with a *problem-solution* relation, but its presence does not guarantee that this is the actual relation between the sentences. Judge the relation from the sentence content itself, not from the connecting phrase alone. Ignore the connecting phrase when assessing the match.

1. Problem (S1)
Return "True" only if:
- S1 states a clear, specific problem (expert terminology an expert recognizes as a problem, OR a descriptive articulation of one). Naming a problem category without saying what makes it a problem is NOT clear and specific.
- Self-contained: the core problem is articulated in S1. A pronoun/reference pointing outside S1 is acceptable only if it is NOT the core problem.
Return "False" if: grammatically incorrect/incoherent; it states a problem exists but not what it is; the core problem is hidden behind a reference pointing outside S1; or S1 isn't actually a problem - it's a capability, requirement, or assumption.

2. Solution (S2)
Return "True" only if:
- S2 describes a solution/mitigation an expert would understand from its terms (established domain terminology counts as sufficiently described).
- Self-contained: the core solution is articulated in S2. A reference to the problem sentence is acceptable; a reference elsewhere is acceptable only if it is NOT the core solution.
Return "False" if: grammatically incorrect/incoherent; it only states a solution exists / gives a bare name / defers details ("the following ...") without saying what it is or does; or S2 only explains how a method works, or is a diagnostic/experimental setup, or is a consequence/derivation, not a remedy.

3. Match
Return "True" only if both S1 and S2 are valid AND S2 directly targets the core problem in S1. Otherwise "False".

OUTPUT (JSON only):
{"is_problem_valid":{"is_valid":"True|False","reason":"<1 sentence>"},"is_solution_valid":{"is_valid":"True|False","reason":"<1 sentence>"},"match_evaluation":{"is_match":"True|False","reason":"<1 sentence>"}}
\end{lstlisting}

\begin{lstlisting}[style=promptstyle,caption={\texttt{Silver Query sentence-Gold sentence: \specrel{} - Prompt 1}.},label={lst:silver_prompts_classifier_specify_p1}]
You judge whether SENTENCE_2 is "more specific" than SENTENCE_1.
Both sentences are drawn from the same scientific paper.

INPUT FORMAT
The input is "S1. <connecting phrase>, S2". The connecting phrase was chosen because it signals this relation, but it only marks the AUTHOR'S INTENT - it does not guarantee the content fits. Judge the actual sentences, not the phrase. A "specifically" / "for example" cue is a hint, not proof.

DEFINITION
S2 is "more specific" when it instantiates, elaborates, or narrows a claim made generally in S1. S2 takes the general statement and supplies a concrete case, a named example, exact figures, an enumerated set of members, or the detailed mechanism that realizes it. The direction runs from general (S1) to particular (S2).

THE CORE TEST (apply in order)
1. Identify the general claim in S1.
2. Ask: is S2 an INSTANCE OF, or a FINER-GRAINED ACCOUNT OF, that exact claim? It must narrow the SAME proposition, not introduce a different one. Valid forms of "more specific":
   - exemplification: S2 gives a named example of S1's general class (e.g. S1 "several studies explored X"; S2 "Smith et al. did X via Y")
   - enumeration: S2 lists the concrete members S1 generalized over (e.g. S1 "we evaluate distinct LLM families"; S2 "GPT, LLaMA, Claude...")
   - quantification: S2 gives the exact numbers behind S1's qualitative claim (e.g. S1 "model improves"; S2 "+2.4% on benchmark Z")
   - mechanism/elaboration: S2 spells out HOW S1's stated operation works (e.g. S1 "tunable embeddings replace [v]"; S2 "'a photo of a [v] dog' becomes 'a photo of a [v_l] dog' per layer")
   - worked case/scenario: S2 grounds S1's principle in one concrete situation.

ANCHOR CHECK
S2's referring expressions ("this", "it", "these results", "the method") must resolve cleanly to S1's content.
If S2's anchor instead points to the paper as a whole ("this work shows...") and is NOT recoverable from S1, the pair is loose - decide false.

DECIDE true (more specific) ONLY IF ALL HOLD:
- S1 states something more general than S2.
- S2 narrows the SAME proposition (instance / detail / enumeration / numbers / mechanism), not a different or sibling proposition.
- Both sentences are individually coherent and not too vague to assess.
- S2's anchor resolves to S1.

DECIDE false IF ANY HOLD:
- S2 is MORE GENERAL than S1 (abstracts it) - that is the inverse relation.
- S1 and S2 are PARALLEL/SIBLING points at the same level: two coordinate effects, two advantages, two future-work directions, a contrast ("however", "in contrast"), or two restatements of one claim.
- ELABORATION/JUSTIFICATION: S2 explains the mechanism or objective rather than giving a narrower case of S1, or S2 justifies why S1 holds rather than narrowing it.
- CO-LEVEL RESULT: S1 is a result/performance claim and S2 is co-level evidence or numbers (same-level result reporting), rather than a general claim that S2 instantiates.
- S2 introduces NEW content (a different mechanism, a contrasting concept, a separate finding) rather than narrowing S1.
- S2 is an INFERENCE/CONSEQUENCE drawn from S1 rather than a concrete instance of it.
- The two sentences are on DIFFERENT TOPICS with no instance relation.
- Either sentence is garbled, truncated mid-clause, or too vague to evaluate - if so, decide false and say so in the reason.

PARTIAL SPECIFICATION
If S1 makes a multi-part claim and S2 instantiates only ONE part, this still counts as true, but note the partial coverage in the reason and lower confidence slightly.

MIXED S2
If S2 begins by specifying S1 but then adds a new/contrasting clause, label on the DOMINANT move; if specification leads, true with a note; if the new material dominates, false.

OUTPUT (JSON only, no preamble, no markdown):
{"decision": true|false, "confidence": <float 0.0-1.0>, "reason": "<one short sentence>"}

\end{lstlisting}

\begin{lstlisting}[style=promptstyle,caption={\texttt{Silver Query sentence-Gold sentence: \specrel{} - Prompt 2}.},label={lst:silver_prompts_classifier_specify_p2}]
You judge whether SENTENCE_2 is "more specific" than SENTENCE_1.

INPUT FORMAT
The input is "S1. <connecting phrase>, S2". The connecting phrase was chosen because it signals this relation, but it only marks the AUTHOR'S
INTENT - it does not guarantee the content fits. Judge the sentences, not the phrase. A "specifically" / "for example" cue is a hint, not proof.

DEFINITION
S2 is "more specific" when it narrows the SAME claim S1 makes generally - via a concrete example, an enumeration of the members S1 generalized over,
exact figures behind a qualitative claim, or the detailed mechanism that realizes it. Direction: general (S1) -> particular (S2).

TEST: is S2 an instance of, or a finer-grained account of, S1's exact claim?

Decide true only if: S1 is more general; S2 narrows the SAME proposition; both sentences are coherent and not too vague; S2's references resolve to S1.

Decide false if: S2 is more GENERAL than S1 (inverse); the two are PARALLEL/sibling points (two effects, two advantages, a contrast, a restatement); S2 adds NEW content or is an inference/consequence rather than an instance; the topics differ; or either sentence is garbled or too vague.

If S1 is multi-part and S2 instantiates only one part, still true - note the partial coverage and lower confidence.
If S2 specifies then adds new material, label on the dominant move.

OUTPUT (JSON only, no preamble or markdown):
{"decision": true|false, "confidence": <float 0.0-1.0>, "reason": "<one short sentence>"}
\end{lstlisting}

\begin{lstlisting}[style=promptstyle,caption={\texttt{Silver Query sentence-Gold sentence: \broadrel{} - Prompt 1}.},label={lst:silver_prompts_classifier_broaden_p1}]
You are evaluating sentence pairs from scientific literature for an information retrieval benchmark.
You are given a single string of the form "S1. connecting term, S2". Determine whether S2 is a *more-general* expansion of S1.

Note: The connecting term is consistent with a *more-general* relation, but its presence does not guarantee that this is the actual relation between the sentences. Judge the relation from the sentence content itself, not from the connecting term alone. Ignore the connecting term when assessing direction and scope.

Definition of the "more-general" relation:
S2 expands on S1 by stating something broader, higher-level, or more general than S1. S1 is the specific; S2 broadens it. Broadening is meant inclusively - S2 counts as more general if it does ANY of the following relative to S1:
- abstracts S1 into a wider principle, category, or claim;
- states the cross-domain applicability of S1 ("not specific to X; any domain where ... can do the same");
- frames S1's specific result/method at the paper level - its broader significance, contribution, or implication;
- draws a broader implication, recommendation, or future direction from S1;
- states a related but higher-level claim that S1's content falls under.
The relation is DIRECTIONAL: S1 is the specific, S2 is the broader/higher-level statement. It does NOT require S1 to be a literal instance of S2's exact proposition - a looser "S1 is an aspect of / contributes to / motivates the broader statement S2" is acceptable, as long as S2 is genuinely broader and connected to S1.

Evaluate the criteria below in order. Return 'true' only if ALL hold:
1. Grammar/coherence: Both S1 and S2 are grammatically correct and coherent on their own.
2. Connection: S2 directly continues from or refers back to S1. If S2 contains references ("this", "these", "such", "our findings", "the principles demonstrated here", "table N", etc.), they must resolve to content present in S1 - not to a prior sentence, a figure/table, or external context.
3. Direction & scope: S2 is broader / higher-level than S1 in any of the inclusive senses above.

Return 'false' if ANY of the following hold (checked in the same order):
- GRAMMAR: Either sentence is grammatically incorrect, garbled, or incoherent.
- CONNECTION: S2 refers to a sentence other than S1, or to figures/tables/findings not present in S1, or its references cannot be resolved from S1 alone.
- INVERSE: S2 is MORE specific than S1 (the inverse relation). A pair may be coherent and well-connected and still fail here.
- SAME LEVEL: S2 merely restates S1 or continues the topic at the same level of specificity without broadening scope.
- UNRELATED: S2 is not topically connected to S1.

Do NOT reject solely because S2 frames things at the paper level ("this work / the approach / the benchmark shows ..."), broadens to other domains, or draws a broader implication - these are valid forms of more-general here, provided S2 is broader than S1 and its references resolve to S1.

Reason field:
- If 'false' because either sentence is grammatically incorrect or incoherent, set "reason" to exactly "grammar_fail".
- Otherwise, give one short sentence naming the failing criterion: inverse / same-level / unrelated / connection.

Output format: Return only a JSON object:
{"decision": true/false, "confidence": [0-1], "reason": "<one short sentence>"}

Input:
{input_string}
\end{lstlisting}

\begin{lstlisting}[style=promptstyle,caption={\texttt{Silver Query sentence-Gold sentence: \broadrel{} - Prompt 2}.},label={lst:silver_prompts_classifier_broaden_p2}]
You are evaluating sentence pairs from scientific literature for an information retrieval benchmark. 
You are given a single string of the form "S1. connecting term, S2". Determine whether S2 is a *more-general* expansion of S1.

Note: The connecting term is consistent with a *more-general* relation, but its presence does not guarantee that this is the actual relation between the sentences. Judge the relation from the sentence content itself, not from the connecting term alone. Ignore the connecting term when assessing direction and scope.

Definition of the "more-general" relation:
S2 expands on S1 by stating something broader, higher-level, abstracting, or more general than S1. That is, S1 expresses a specific case, instance, or detail, and S2 generalizes from it to a wider principle, category, or claim. The relation is DIRECTIONAL: S1 is the specific, S2 is the general.

Evaluate the criteria below in order. Return 'true' only if ALL hold:
1. Grammar/coherence: Both S1 and S2 are grammatically correct and coherent on their own.
2. Connection: S2 directly continues from or refers back to S1. If S2 contains references ("this", "these", "such", "the above", etc.), they must resolve to content present in S1 - not to a prior sentence or external context.
3. Direction & scope: S2 is strictly broader in scope than S1. S1 must read as the specific case and S2 as the generalization of it.

Return 'false' if ANY of the following hold (checked in the same order):
- GRAMMAR: Either sentence is grammatically incorrect, garbled, or incoherent.
- CONNECTION: S2 refers to a sentence other than S1, or its references cannot be resolved from S1 alone (assumes missing context).
- INVERSE: S2 is MORE specific than S1 (this is the inverse relation, not more-general). A pair may be coherent and well-connected and still fail here.
- SAME LEVEL: S2 merely continues the topic at the same level of specificity without broadening scope.
- UNRELATED: S2 is not topically connected to S1.
- Other: Any other reason that prevents S2 from being a more-general expansion of S1.

Note: criteria 2 (connection) and 3 (direction) are independent. A pair can be correctly connected via anaphora yet still be same-level or inverse - such pairs are 'false'.

Reason field:
- If the decision is 'false' because either sentence is grammatically incorrect or incoherent, set "reason" to exactly "grammar_fail".
- Otherwise, give one short sentence explaining the decision (e.g., naming the failing criterion: inverse / same-level / unrelated / connection).

Output format: Return only a JSON object:
{"decision": true/false, "confidence": [0-1], "reason": "<one short sentence>"}

Input:
{input_string}
\end{lstlisting}

\subsection{Reranker}
\label{app:reranker-prompt}
The system message is the instruction below with the relation line filled in per operation; the user message is \texttt{Query: <query>} followed by the numbered candidates.

\begin{lstlisting}[style=promptstyle,caption={\texttt{Reranker: system message}.},label={lst:reranker_prompt}]
You rerank candidate sentences taken from computer-science papers for one query sentence.
Target relation ("{move}"): {relation}

Rules:
- Judge each candidate only by how well it stands in the target relation to the query. Topical similarity alone is not enough; a candidate that merely restates the query does not satisfy the relation.
- The candidates are numbered [1]..[N]. Their order carries no information about relevance.
- Return the indices of the 10 candidates that best satisfy the relation, best first. Use only indices that appear in the list, no duplicates.
- Output JSON only: {"ranking": [<index>, ...]}
\end{lstlisting}

\noindent\texttt{\{relation\}} is, per operation:
\begin{itemize}
\item \addrrel{}: \textit{the candidate proposes, describes or evaluates a solution, improvement or mitigation for the problem, limitation or need stated in the query.}
\item \broadrel{}: \textit{the candidate states the query's research problem, idea or method in a MORE GENERAL form: the query is a specific instance or special case of what the candidate describes.}
\item \specrel{}: \textit{the candidate states a MORE SPECIFIC instance, special case or concrete realisation of the research problem, idea or method described in the query.}
\end{itemize}

\end{document}